\documentclass[conference]{IEEEtran}
\IEEEoverridecommandlockouts

\usepackage{cite}
\usepackage[thinc]{esdiff}  
\usepackage{amsmath,amssymb,amsfonts,amsthm,bm}
\usepackage{commath}  
\usepackage{graphicx}
\usepackage{textcomp}
\usepackage{xcolor}
\usepackage{algorithm}
\usepackage[noend]{algpseudocode}
\usepackage[nocomma]{optidef}  
\usepackage{subcaption} 
\usepackage[thinc]{esdiff}  

 \usepackage[table, dvipsnames]{xcolor}
\usepackage{adjustbox}
\usepackage{multicol}
\usepackage{tabularx}
\usepackage{caption}
\usepackage{amssymb}
\usepackage{cellspace}
\usepackage{hyperref}
\usepackage{cite}
\usepackage{multirow}
\usepackage{float}
\usepackage{booktabs}
\usepackage{wrapfig}

\usepackage[font=small,labelfont=bf]{caption}

\newcommand{\B}{\mathbf}
\newcommand{\C}{\mathcal}

\makeatletter
\algnewcommand{\LineComment}[1]{\Statex \hskip\ALG@thistlm \(\triangleright\) #1}
\algnewcommand{\LineCommentCont}[1]{\Statex \hskip\ALG@thistlm \parbox[t]{\linegoal}{\hangindent=1em\hangafter=1 $\triangleright$ #1}}
\makeatother

\algnewcommand\algorithmicforeach{\textbf{for each}}
\algdef{S}[FOR]{ForEach}[1]{\algorithmicforeach\ #1\ \algorithmicdo}

\definecolor{mygray}{gray}{0.6}
\definecolor{mypink}{cmyk}{0, 0.7808, 0.4429, 0.1412}
\def\BibTeX{{\rm B\kern-.05em{\sc i\kern-.025em b}\kern-.08em
    T\kern-.1667em\lower.7ex\hbox{E}\kern-.125emX}}
\begin{document}

\title{\vspace{+1.5em}SPOT: Spatio-Temporal Obstacle-free Trajectory Planning for UAVs in unknown dynamic environments}
\author{Astik Srivastava, Thomas J Chackenkulam, Bitla Bhanu Teja, Antony Thomas, and Madhava Krishna
\thanks{Astik Srivastava, Bitla Bhanu Teja, Antony Thomas and Madhava Krishna are with Robotics Research Center, IIIT Hyderabad, India. Email:(astik.srivastava, bhanu.teja)@research.iiit.ac.in, (antony.thomas, , mkrishna)@iiit.ac.in}
\thanks{Thomas J Chackenkulam is with IIT-BHU, India. Email: thomasj.chackenkulam.cd.mec23@itbhu.ac.in}
}

\maketitle

\begin{abstract}
We address the problem of reactive motion planning for quadrotors operating in unknown environments with dynamic obstacles. Our approach leverages a 4-dimensional spatio-temporal planner, integrated with vision-based Safe Flight Corridor (SFC) generation and trajectory optimization. Unlike prior methods that rely on map fusion, our framework is mapless, enabling collision avoidance directly from perception while reducing computational overhead. Dynamic obstacles are detected and tracked using a vision-based object segmentation and tracking pipeline, allowing robust classification of static versus dynamic elements in the scene. To further enhance robustness, we introduce a backup planning module that reactively avoids dynamic obstacles when no direct path to the goal is available, mitigating the risk of collisions during deadlock situations. We validate our method extensively in both simulation and real-world hardware experiments, and benchmark it against state-of-the-art approaches, showing significant advantages for reactive UAV navigation in dynamic, unknown environments.
\end{abstract}

\section{Introduction}
\label{sec:introduction}
Quadrotors have emerged as the preferred unmanned aerial vehicle (UAV) platform for both civilian and military applications due to their simple mechanical design and ease of control. This has led to their widespread adoption in diverse domains such as aerial photography, surveillance, surveying, and disaster response. However, achieving safe and autonomous navigation in complex and dynamic environments remains a critical research challenge~\cite{tordesillas2021TRO,lu2025TRO}.

A widely adopted approach to quadrotor navigation is a two-stage planning framework~\cite{gao2020TRO, faster, wang2022TRO, ren2022bubble, ren2025super}. In this framework, a \textit{global} planner first computes a feasible path, which serves as the foundation for collision free polyhedral regions. A \textit{local} planner then dynamically generates smooth and feasible trajectories within these regions, ensuring real-time adaptability while maintaining safety and efficiency. While these approaches perform well in environments with static obstacles, their performance degrades significantly when faced with dynamic obstacles.

\begin{figure}
    \begin{subfigure}[b]{0.47\linewidth}
        \centering
        \includegraphics[width=\linewidth]{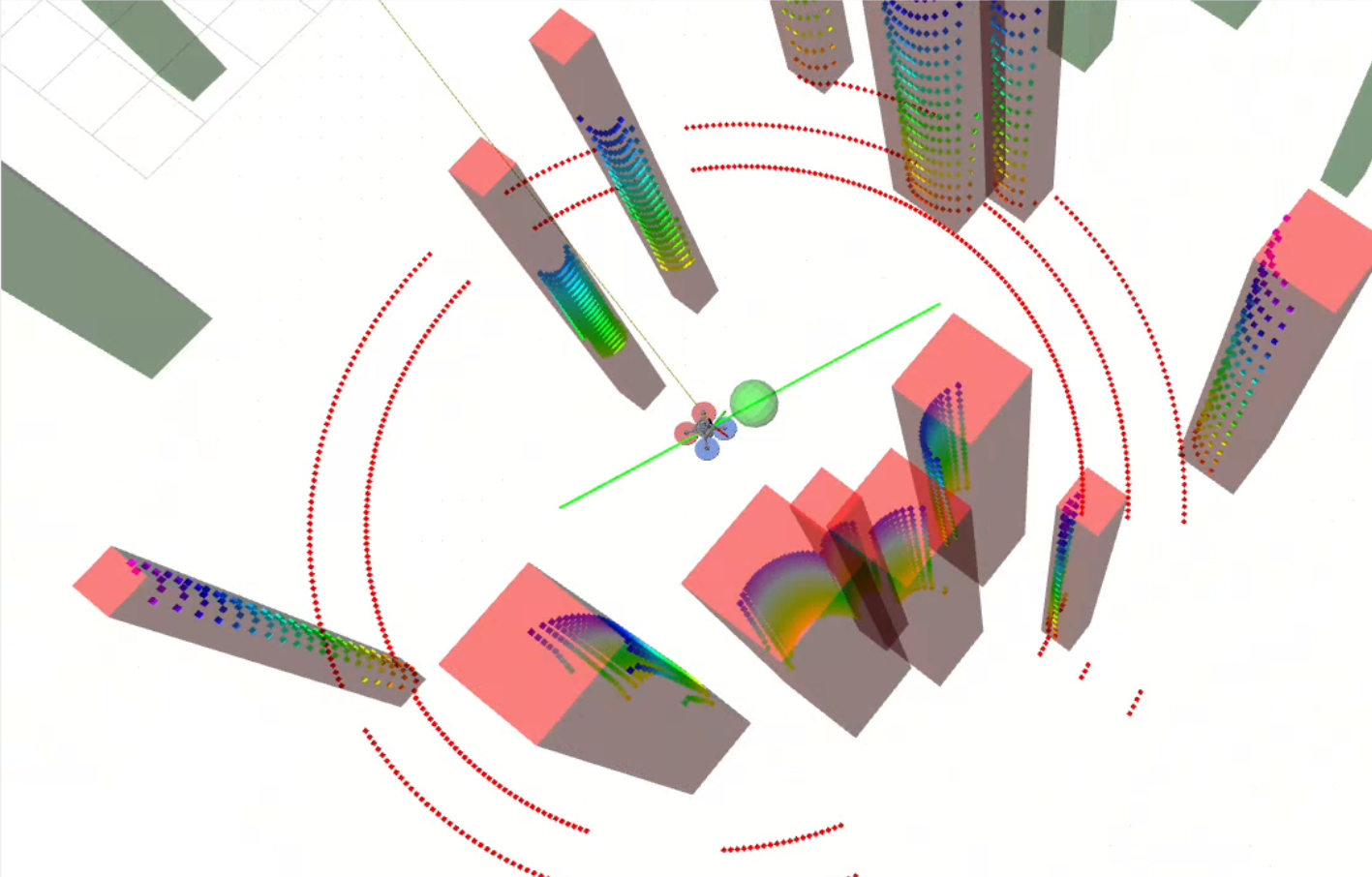}
        \caption{Deadlock encounter}
    \end{subfigure}
    \hfill
    \begin{subfigure}[b]{0.47\linewidth}
        \centering
        \includegraphics[width=\linewidth]{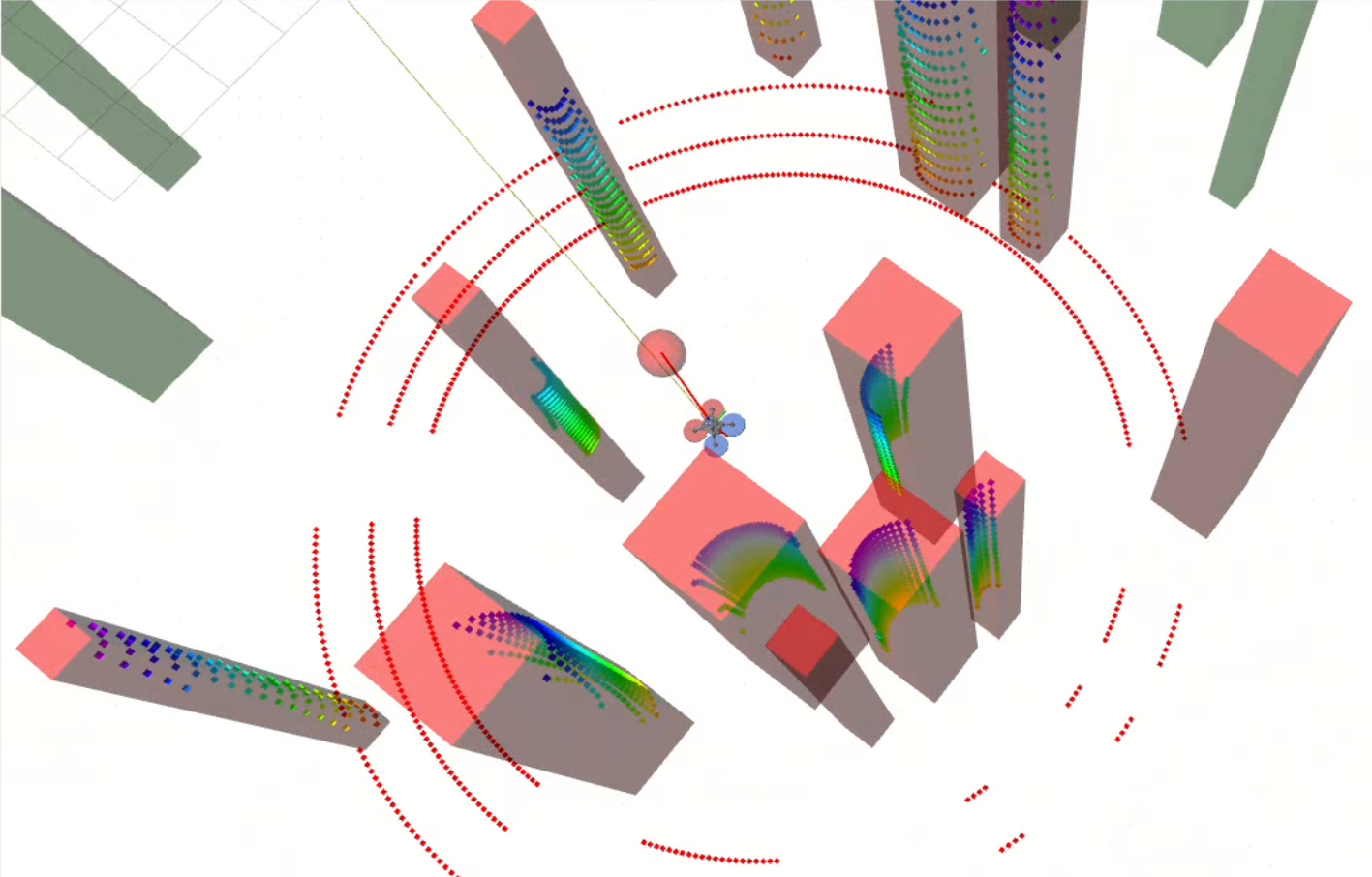}
        \caption{Backup Generation}
    \end{subfigure}
    \begin{subfigure}[b]{0.47\linewidth}
        \centering
        \includegraphics[width=\linewidth]{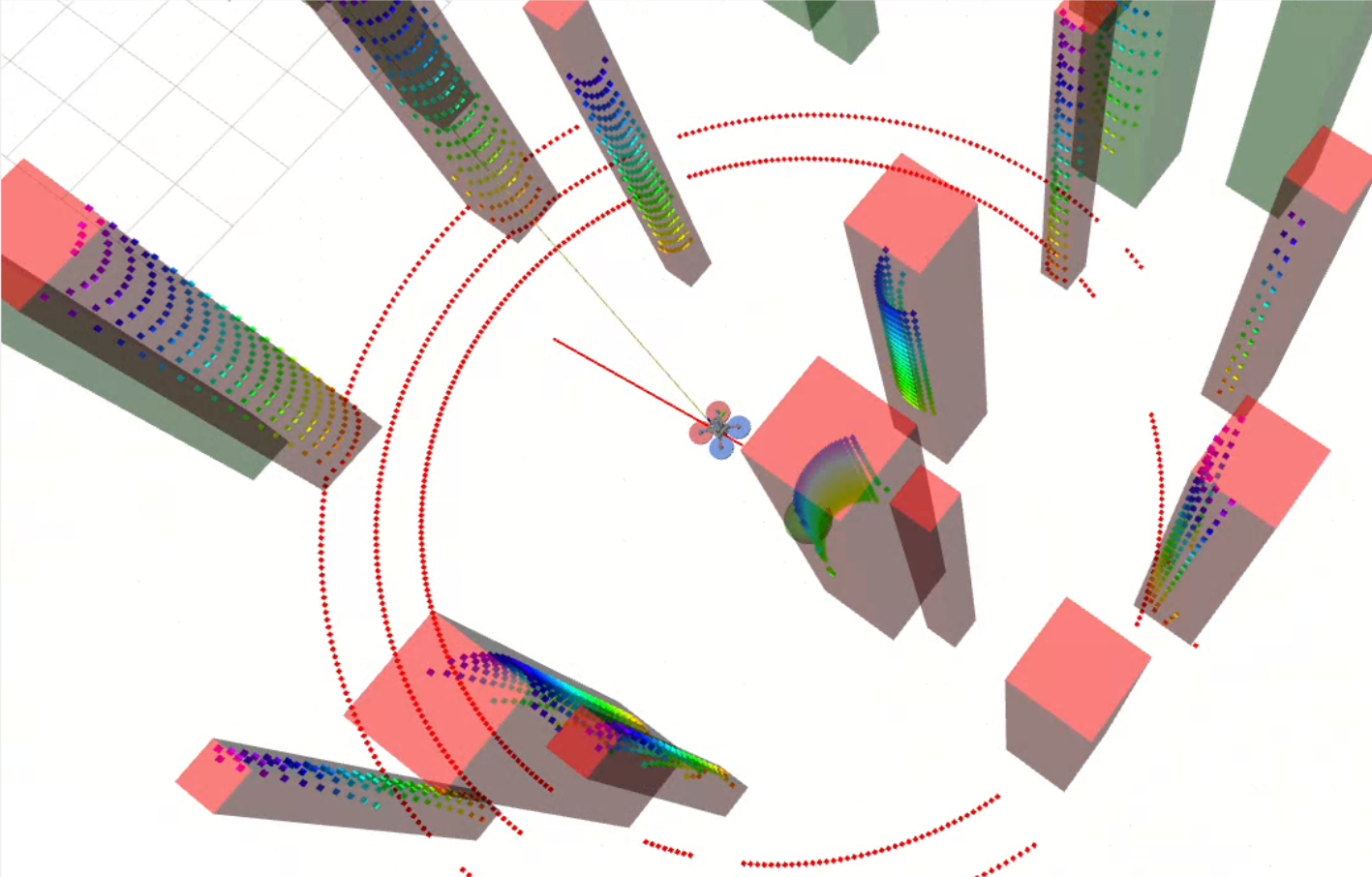}
        \caption{Backup Tracking}
    \end{subfigure}
    \hfill
    \begin{subfigure}[b]{0.47\linewidth}
        \centering
        \includegraphics[width=\linewidth]{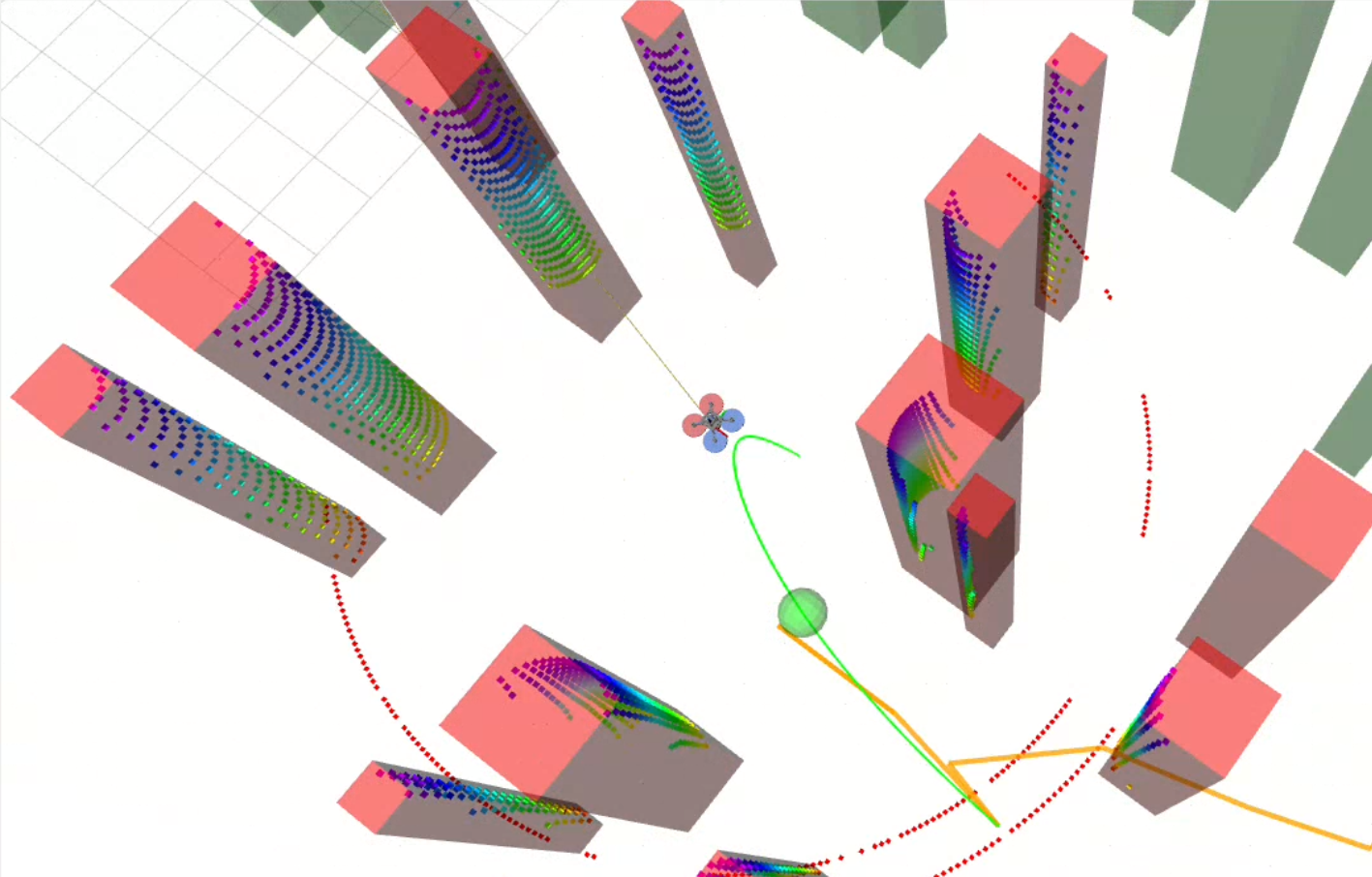}
        \caption{Replanning}
    \end{subfigure}
    \caption{UAV navigation in a dynamic environment with backup planning. (a) The nominal trajectory (green) becomes unsafe due to dynamic obstacles. (b) A backup trajectory (red) is generated reactively to move the UAV toward a safer region. (c) The UAV tracks the backup trajectory, replanning it as needed to maintain safety. (d) Once a feasible path to the goal becomes available, the UAV switches back to goal-directed planning. Convex collision-free corridors are omitted for clarity. Obstacles within the UAV’s sensing range are shown with red bounding boxes, while those outside the range are shown with green bounding boxes.}
    \label{teaser}
\end{figure}

Recently, several methods have been proposed to address motion planning with dynamic obstacles~\cite{lin2020ICRA,he2021IROS,xu2022ICRA,hou2022RAL,lu2022RAL,RAST,wu2024ICRA,lu2025TRO,quan2025RAL}. Despite this progress, existing approaches remain limited in several key aspects. Many works~\cite{lin2020ICRA,he2021IROS,lu2022RAL,xu2022ICRA} consider only a handful of dynamic obstacles, which is far from the complexity encountered in realistic, cluttered environments. Some methods rely on motion capture systems to provide ground-truth obstacle trajectories~\cite{he2021IROS, RAST, wu2024ICRA,lu2025TRO, quan2025RAL}, an assumption that is impractical for real-world deployment. Furthermore, most approaches~\cite{lin2020ICRA,he2021IROS,xu2022ICRA,hou2022RAL,lu2022RAL,RAST,wu2024ICRA,quan2025RAL} fail to incorporate backup or contingency trajectories. This omission is critical, as in densely populated environments with multiple dynamic obstacles, trajectory optimization alone often fails, leaving the robot without safe fallback (see Fig.~\ref{teaser}). Overcoming these limitations is therefore essential for achieving robust and reliable motion planning in dynamic, real-world settings.

In this work, we propose Spatio-Temporal Obstacle-free Trajectory (SPOT), a framework for autonomous navigation of UAVs in unknown dynamic environments, designed to address the challenges outlined above. Specifically, our key contributions are as follows
\begin{itemize}
    \item A novel formulation of spatio-temporal RRT$^\star$ motion planning for mapless, vision-based navigation in dynamic environments.
    \item A robust planning pipeline that explicitly handles deadlocks and prevents collisions through backup trajectory generation.
    \item Extensive validation in both simulation and real-world hardware, demonstrating reliable performance using only onboard sensing and computation. Simulations include scenarios with up to 30 dynamic obstacles. For reproducibility, the code is made available\footnote{\url{https://astik-2002.github.io/ICRA-2026-SPOT/}}.

\end{itemize}



\section{Related Work}
\label{sec:related_work}
\subsection{Global planning and safe flight corridor}
A well-established approach to collision avoidance in autonomous navigation involves performing convex decomposition in the configuration space based on the results of a global planner, followed by local trajectory optimization within the decomposed convex corridors~\cite{deits2015computing, liu2017planning, faster, gao2019flying, ren2022bubble,marcucci2023SR,wang2024fast,ren2025super}. Convex decomposition is performed around the piecewise-linear path generated by the global planner. Each segment of this path is enclosed within a polyhedron, constructed by first inflating an ellipsoid along the segment and then computing tangent planes at the contact points between the ellipsoid and obstacles. The collection of such overlapping convex polyhedra forms a Safe Flight Corridor (SFC)~\cite{liu2017planning}, which defines a collision-free region for trajectory generation.

Global planning algorithms typically employed in this framework include grid-based methods (e.g., Jump Point Search, A$^\star$)~\cite{ liu2017planning, faster, wu2024ICRA}, as well as sampling-based approaches (e.g., RRT and its variants)~\cite{richter2016ISRR}. However, state-of-the-art SFC-based methods~\cite{faster, ren2025super} largely remain restricted to static obstacles in unknown environments, limiting their applicability in dynamic settings.
\subsection{UAV planning in dynamic environments}
A number of recent methods address trajectory planning for UAVs in dynamic environments. One line of work first constructs SFC by considering only static obstacles, and subsequently enforces dynamic collision-avoidance constraints during trajectory optimization. Examples include chance-constrained MPC formulations~\cite{lin2020ICRA,xu2022ICRA,hou2022RAL}, where dynamic obstacles are typically modeled using simplified geometric primitives such as ellipsoids, spheres, or bounding boxes~\cite{kamel2017IROS,lin2020ICRA,tordesillas2021TRO}. An alternative strategy proposed in~\cite{tordesillas2022ACCESS} builds convex hulls around individual dynamic obstacles; however, this approach becomes computationally inefficient as the number of obstacles increases. Lu \textit{et al.} deals with only small moving obstacles such as tennis balls.

In contrast, other works model static and dynamic obstacles jointly~\cite{chen2023TIE,RAST,lu2025TRO,quan2025RAL}. Chen \textit{et al.}~\cite{chen2023TIE} propose a dual-structure particle-based map to simultaneously capture both obstacle types. Similarly, RAST~\cite{RAST} employs a particle-based occupancy map combined with sampling-based planning to generate trajectories that minimize particle collisions. However, RAST restricts SFC construction to cuboid-shaped corridors, which can be overly conservative in environments with irregular geometries. In~\cite{lu2025TRO}, static and dynamic obstacles are jointly formulated as collision-avoidance constraints within the trajectory optimization process. More recently, Quan \textit{et al.}~\cite{quan2025RAL} utilize a state-time space representation derived from the Euclidean Signed Distance Field (ESDF) to predict pedestrian motion, though their formulation remains limited to 2D environments.

In this work, we construct a spatio-temporal SFC that simultaneously accounts for static and dynamic obstacles. A key distinction is that our framework explicitly incorporates backup trajectories, which are critical in cluttered, dynamic environments where the UAV may otherwise encounter deadlock due to obstacles obstructing its forward motion. Existing approaches~\cite{chen2023TIE,RAST,faster,wang2024fast,quan2025RAL,ren2025super} do not address this issue. The work of Lu \textit{et al.}~\cite{lu2025TRO} is arguably the closest, as it also considers fallback strategies, but their formulation relies on maintaining a static local map. In contrast, our approach (SPOT) is map-less, making it light-weight and more scalable to highly dynamic and unknown environments.


\section{Methodology}
\label{sec:methodology}
\begin{figure}
    \centering
    \includegraphics[width=0.8\linewidth]{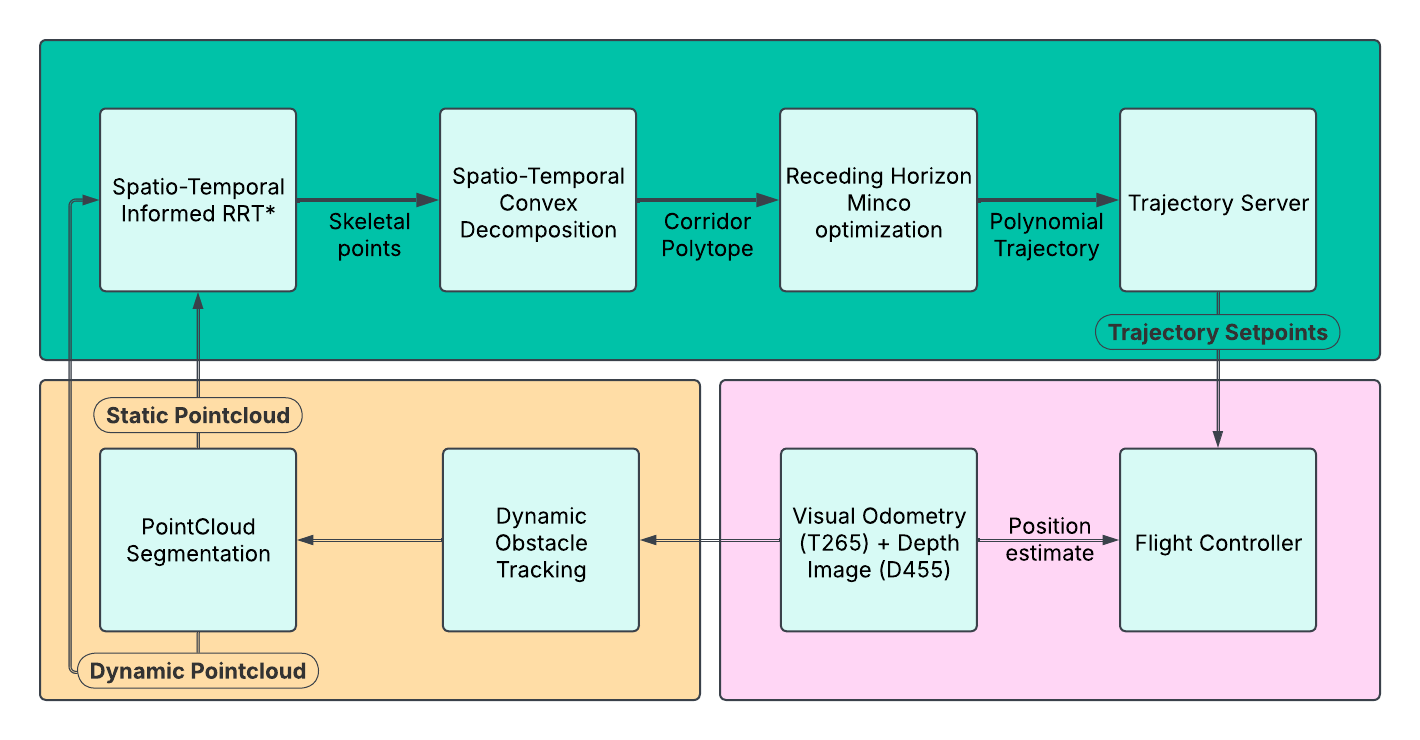}
    \caption{System architecture of SPOT for reactive UAV motion planning in dynamic, unknown environments.}
    \label{system_architecture}
\end{figure}
\subsection{Planning pipeline}
\begin{figure}
    \centering
    \includegraphics[width=0.7\linewidth]{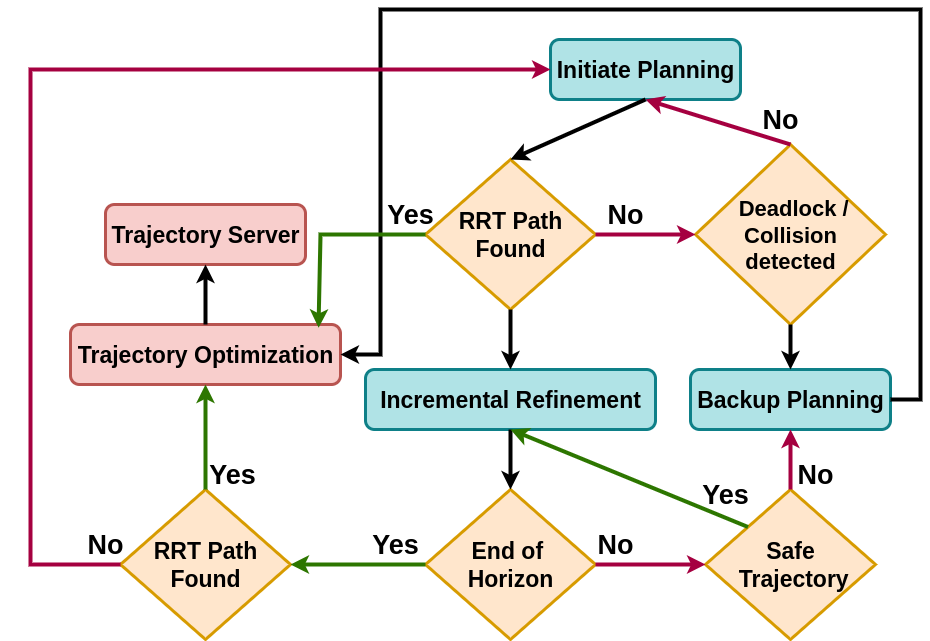}
    \caption{FSM governing the spatio-temporal RRT$^\star$ planning pipeline, operating in three modes: \textit{Initial}, \textit{Incremental}, and \textit{Backup}.}
    \label{fig_architecture}
\end{figure}

In this section, we present our planning architecture for dynamic collision avoidance. An overview of the system is shown in Fig.~\ref{system_architecture}. The proposed framework SPOT consists of three main components:
\begin{itemize}
    \item \textbf{Object Motion Segmentation}: Since planning is performed directly on point clouds (obtained from the onboard depth camera), it is critical to estimate both the position and velocity of points in the scene. To this end, we employ the onboard dynamic object tracking framework~\cite{10323166}, which outputs bounding boxes along with object positions and velocities. Each point lying inside a bounding box inherits the corresponding object velocity, while all other points are treated as static. This step produces a spatio-temporal representation of the environment.
    \item \textbf{Collision Free Skeletal Path Generation}: Using the classified point cloud, we generate a guiding path to the goal via spatio-temporal RRT$^\star$ (see Section~\ref{subsec:st_rrt}). Unexplored regions outside the sensor’s field of view (FOV) are assumed obstacle-free, consistent with the mapless design. Path generation is regulated by a Finite State Machine (FSM), which serves as the decision-making layer. An overview of the same is given in Fig.~\ref{fig_architecture} and discussed subsequently. The FSM determines whether to initiate a new search, refine an existing path, or invoke the backup planner, thereby ensuring safe and reliable UAV navigation under dynamic conditions.
    \item \textbf{Convex Decomposition and Trajectory Generation}: The skeletal path is then inflated into a safe corridor using convex decomposition, after which a smooth, dynamically feasible trajectory is generated. This ensures collision-free execution while adhering to UAV dynamics. 
\end{itemize}

The planning pipeline is governed by an FSM-based architecture with three modes: \textit{Initial}, \textit{Incremental}, and \textit{Backup}. 
The initial and incremental planning blocks build upon the framework in \cite{gao2019flying}, with modifications to support efficient spatio-temporal planning. In the initial mode, RRT$^\star$ is invoked to plan a path from the UAV’s starting state to the goal. Once a valid path is found, the portion of the tree within the sensing horizon is used for SFC generation and trajectory optimization. In the incremental mode, the path is continuously refined by adding new nodes to the tree while simultaneously evaluating the safety of the trajectory currently being executed by the UAV. Finally, in the backup mode, if the planner fails to find a feasible path within the time budget, or if the UAV is hovering at a potentially unsafe location, a reactive backup planner is activated to generate an immediate collision-free maneuver, thereby reducing the risk of collisions with dynamic obstacles.
\subsection{Spatio-Temporal RRT}
\label{subsec:st_rrt}

\begin{figure}
    \centering
    \includegraphics[width=0.7\linewidth]{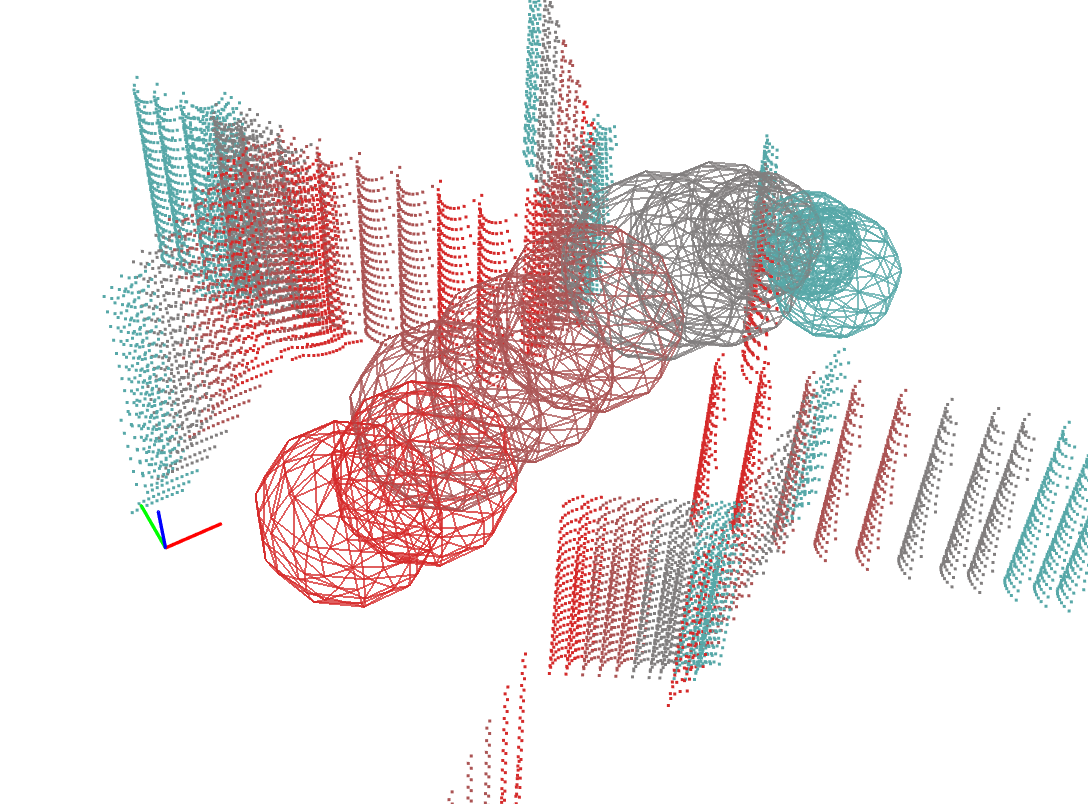}
    \caption{Spatio-temporal RRT path generated in a dynamic environment. The environment provides a point cloud of obstacles, propagated into the future using velocity information. Each sphere represents the safety margin of a node (feasibility set) at its time of arrival. A color gradient from red to blue illustrates the flow of time, from the present (red) toward the future (blue).}
    \label{fig:st-rrt}
\end{figure}

In this section, we present our Spatio-Temporal RRT$^{\star}$ (ST-RRT$^\star$) algorithm for collision avoidance. In a dynamic environment the occupancy of any given location inherently time-dependent. In other words, whether a location is occupied depends not only on the obstacle’s trajectory but also on the arrival time of the robot at that location. To account for this, we augment the standard RRT$^\star$ presented in \cite{gao2019flying} by propagating the robot’s travel time from the \textit{root} node to every node in the tree. Formally, each node is represented as
$$
n_i = \left(x_i, y_i, z_i, t_i\right)
$$
where $(x_i, y_i, z_i)$ denotes the position in 3D space and $t_i$ is the accumulated travel time from the root node. This representation allows each node to encode both spatial position and arrival time, thereby enabling collision checks to be performed consistently against obstacle trajectories in the spatio-temporal space. An illustration is provided in Fig.~\ref{fig:st-rrt}.

\subsubsection{Travel time estimate}
To incorporate the arrival time at each node, we employ a kinematics-based travel-time estimate, which is propagated and encoded into every node during the ST-RRT$^\star$ construction. Let $v_{max}$ denote the maximum admissible velocity for the robot. Consider a parent node $n_i= (\B{p}_i, t_i), \B{p}_i = [x_i, y_i, z_i]^T$ and a child node $n_j= (\B{p}_j, t_j), \B{p}_j = [x_j, y_j, z_j]^T$. The constructed spatio-temporal tree must satisfy two conditions: (i) causality, that is, the arrival time at any child node must strictly exceed that of its parent, and (ii) velocity feasibility, that is, the child node must be reachable from its parent under the maximum velocity bound $\B{v}_{max}$ (see Fig.~\ref{fig:st-rrt}). Therefore, a child node is admissible if and only if it belongs the feasibility set
\begin{equation}
\label{consistent_nodes}
    \C{F}(n_i) = \Big\{n_j \ \big| \ t_j > t_i, \; \ \norm{\B{p}_j-\B{p}_i} \leq \B{v}_{max}(t_j-t_i)\Big\}
\end{equation}
This formulation ensures that each edge in the tree is both temporally ordered and kinematically feasible, thereby embedding spatio-temporal consistency directly into the RRT$^\star$ expansion process. Moreover, the feasibility set $\C{F}(n_i)$ naturally extends to incorporate additional kinematic constraints, such as bounded acceleration.




\subsubsection{Edge cost}
Given the spatio-temporal nature of the planning problem, we define a heuristic cost function for the connection between a parent node and a child node as
\begin{equation}
    C_e = \, \ \norm{\B{p}_j - \B{p}_i}+ w \ (\,t_j - t_i)
\end{equation}
\noindent where $w$ is a weighting parameter which can be interpreted as a \textit{desired velocity}. The above formulation enables a tunable trade-off between spatial distance and temporal separation. This flexibility allows the planner to adapt to application-specific priorities, such as minimizing travel distance or performing cautious motion planning in cluttered  dynamic environments. In the steer function for ST-RRT$^{*}$ and also during rewiring, we ensure that the parent and child nodes are consistent as defined in~\eqref{consistent_nodes} by restricting the connections of a node to other nodes which are reachible given UAV's velocity constraints.
\subsubsection{Sampling strategy}
We have adopt informed sampling\footnote{Informed RRT$^\star$ operates like standard RRT$^\star$ until the first solution is found, after which it restricts sampling to a heuristic-defined subset of states, enabling solution refinement while implicitly balancing exploration and exploitation without additional parameters~\cite{gammell2014IROS}.} to accelerate convergence to an optimal path. Unlike prior work, we extend informed sampling to include not only spatial coordinates but also the time of arrival assigned to each newly added node. This restricts exploration to a spatio-temporal subset of the configuration space guaranteed to contain any better solution.

Formally, let $c_\text{best}$ be the cost of the best path found from start $\B{p}_s$ to goal $\B{p}_g$. The informed set $\C{X}$ consists of all points $\B{p}$ satisfying:
\begin{equation}
\norm{\B{p}_s - \B{p}} \ + \ \norm{\B{p} - \B{p}_g} \ \leq \ c_\text{best}
\end{equation}
The set $\C{X}$ forms a prolate hyperspheroid with foci at $\B{p}_s$ and $\B{p}g$, and major axis length $c_\text{best}$. Samples are efficiently drawn from this ellipsoid using the closed-form method in \cite{gao2019flying}.

To extend this strategy to spatio-temporal planning, we employ a \textit{decoupled sampling approach}. A candidate spatial sample $\B{p}_\text{rand}$ is first generated within the informed ellipsoid. A time coordinate $t_\text{rand}$ is then assigned to this point by sampling from a distribution designed to prioritize time-efficient solutions
\begin{equation}
t_\text{rand} = t_s + k \ \frac{c_\text{best}}{w}
\end{equation}
where $k \sim \C{U}(0, 1)$ is uniformly distributed, $t_s$ is the start time, and $w$ is the temporal weight from our cost function. Intuitively, this strategy ensures that sampling is focused only on regions of space and time where an improved path could realistically exist. Instead of wasting samples in irrelevant areas or times, the planner concentrates effort within a tightly bounded spatio-temporal subset, thereby accelerating convergence to high-quality, time-efficient trajectories.
\subsubsection{Collision checking}
Collision checking in spatio-temporal planning is particularly challenging, as the safety of a node $n_i$ depends on its distance to obstacles at the specific time $t_i$ it is visited. For static environments, we use a kdtree constructed from all the stationary points, which enables efficient nearest-neighbor queries in static point clouds. However, this approach is not directly applicable in dynamic scenes, since obstacle positions and velocities are only known at the instant sensor measurements are received. Beyond that instant, obstacle motion must be propagated forward in time to estimate occupancy at $t_i$ for safety evaluation. To achieve this efficiently, we employ a spatio-temporal hash grid to perform approximate nearest-neighbor queries. The procedures for hash grid construction and dynamic collision checking are described in Algorithm~\ref{alg:temporal_grid} and Algorithm~\ref{alg:query_dynamic}, respectively. 

Each dynamic point is encoded in the spatio-temporal hash grid by accounting for the corresponding resolutions in time and space (lines 5–7, Algorithm~\ref{alg:temporal_grid}). The resulting hash grid $\C{G}$ is then queried by Algorithm~\ref{alg:temporal_grid} during collision checking. Specifically, the temporal component of a query point is first used to find the nearest temporal bin of the dynamic obstacle by calling the subroutine \texttt{NearestBin} (line 2, Algorithm~\ref{alg:query_dynamic}). The spatial component of the query point is then used to construct all 27 neighbors in 3D (lines 3–6, Algorithm~\ref{alg:query_dynamic}). Each neighbor is checked for occupancy in $\C{G}$, yielding the closest obstacle point (line 11, Algorithm~\ref{alg:query_dynamic}).

The safety margin of a node is defined as the minimum of its static clearance (computed from the kd-tree) and its dynamic clearance $d_{min}$ (obtained from the spatio-temporal hash grid). A node is considered unsafe if this margin is less than the prescribed safety threshold $r_{min}$.

\begin{algorithm}[t]
\caption{Spatio-Temporal Hash Grid}
\label{alg:temporal_grid}
\begin{algorithmic}[1]
\Require{$\C{D}$, time resolution $\Delta t$, cell size $c$, start time $t_{\text{start}}$, end time $t_{\text{end}}$}
\LineComment \textcolor{mypink}{$\C{D}$ is defined as the set of dynamic points of the form $(p, v)$, where $p$ denotes the position and $v$ denotes the associated velocity.}
\Ensure{Temporal grid $\C{G}$}
\State{$\C{G} \gets \emptyset$}
\ForEach{$(p, v) \in \C{D}$}
\For{$t = t_{\text{start}} \ \text{to} \ t_{\text{end}}; \ t = t + \Delta t$}
\State{$p(t) \gets p + v \cdot (t - t_{\text{start}})$}
\State{$t_{\text{bin}} \gets \lfloor t / \Delta t \rfloor$}
\State{$(x_i, y_i, z_i) \gets (\lfloor p_x / c \rfloor, \lfloor p_y / c \rfloor, \lfloor p_z / c \rfloor)$}
\State{$\C{G}[t_{\text{bin}}][(x_i, y_i, z_i)] \gets  p(t)$} 
 \EndFor
 \EndFor
\end{algorithmic}
\end{algorithm}


\begin{algorithm}[t]
\caption{Proximity Check to Nearest Obstacle}
\label{alg:query_dynamic}
\begin{algorithmic}[1]
\Require{Query point $q = (q_x, q_y, q_z,q_t)$, temporal grid $\C{G}$, cell size $c$, time resolution $\Delta t$}
\Ensure{$d_{\min}$}
\LineComment \textcolor{mypink}{$d_{\min}$ represents the minimum distance between the query point and its closest obstacle.}
\State{$d_{\min} \gets \infty, \ t \gets q_t, \ p \gets (q_x,q_y,q_z)$}
\State{$t_\text{bin} \gets$ \texttt{NearestBin}($\C{G}$, t)}
\LineComment \textcolor{mypink}{The subroutine returns the nearest temporal index in $\C{G}$ corresponding to $t$.}
\State{$(x_i, y_i, z_i) \gets (\lfloor p_x / c \rfloor, \lfloor p_y / c \rfloor, \lfloor p_z / c \rfloor)$}
\For{$dx = -1 \ \text{to} \ 1$}
\For{$dy = -1 \ \text{to} \ 1$}
\For{$dz = -1 \ \text{to} \ 1$}
\State{$key \gets (x_i + dx, y_i + dy, z_i + dz)$}
\If{$key \in \C{G}[t_{\text{bin}}]$}
 \ForEach{obstacle $o$ in $\C{G}[t_{\text{bin}}][key]$}
 \State{$d \gets \|p_q - p\|$}
 \State{$d_{\min} \gets \min(d_{\min}, d)$}
\EndFor
\EndIf
\EndFor
\EndFor
\EndFor
\Return{$d_{\min}$}
\end{algorithmic}
\end{algorithm}

\subsection{Spatio-Temporal SFC}
As discussed earlier, SFC is defined as a sequence of overlapping convex polyhedra. We use the configuration space convex decomposition presented in \cite{ren2025super} for finding spatial corridors. Each polyhedron is constructed around a line segment connecting two points in the free space. Since we employ ST-RRT$^\star$ to generate the initial skeletal path, these points correspond to nodes in the path. Each node is associated with a timestamp, enabling spatio-temporal reasoning. Consider two nodes $n_i$ and $n_j$; conventionally, a polyhedron is constructed around the line segment $\overline{n_in_j}$. This is achieved by inflating an ellipsoid along the segment and computing tangent planes at its contact points with obstacles, which are represented as point clouds.

\begin{figure}
    \centering
    \includegraphics[width=0.9\linewidth]{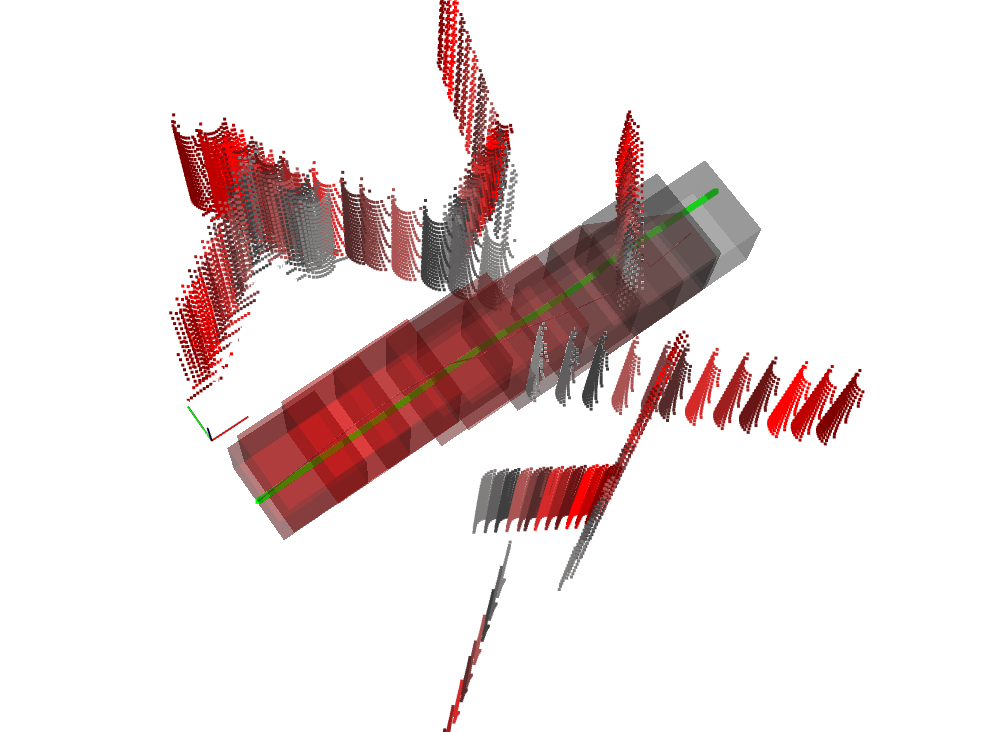}
    \caption{Spatio-temporal convex collision-free polyhedra constructed along the path segment between nodes $n_i$ and $n_j$. The evolution of the safe corridor is color-coded from red to gray over the time span $[t_i, t_f]$. During the same interval, the predicted positions of dynamic obstacles are represented as point clouds, also color-coded from $t_i$ to $t_f$. These aggregated point clouds constitute the dynamic obstacle set $\C{O}^d_{ij}$ used in corridor construction. Final optimized trajectory is visible in green.}
    \label{fig:sfc_dynamic}
\end{figure}

To extend this formulation to dynamic environments, we discretize the temporal interval between $n_i$ and $n_j$ using a step size $\Delta t$, resulting in the set $\C{T}_{ij} =\{t_i, t_i + \Delta t, t_i + 2\Delta t, \ldots, t_j \}$. At each temporal sample $t \in \C{T}_{ij}$, the positions of the dynamic obstacles are predicted under a constant velocity model 
\begin{equation}
    \B{o}(t) = \B{o}(t_i) + \B{v}_o (t- t_i)
\end{equation}
\noindent where $\B{o}(t)$ is the predicted obstacle location based on the observed obstacle position $\B{o}(t_i)$ at time $t_i$ and $\B{v}_o$ is the velocity of the obstacle. The corresponding point clouds are aggregated to form the dynamic obstacle set $\C{O}^{d}_{ij}$. The effective obstacle set to be considered during corridor construction is given by
\begin{equation}
    \C{O}_{ij} = \C{O}^s \cup \C{O}^{d}_{ij}
\end{equation}
where $\C{O}^s$ denotes the static obstacles. In this manner, the SFC construction inherently accounts for both static and dynamic obstacles across the relevant time horizon. The construction of spatio-temporal collision-free polyhedra and the corresponding evolution of dynamic obstacle point clouds are illustrated in Fig.~\ref{fig:sfc_dynamic}. In our implementation, we empirically fix the temporal resolution to $\Delta t = 0.2 \,\text{s}$, while constraining the maximum interval $(t_j - t_i)$ to 2 seconds. These choices strike a balance between prediction fidelity and computational tractability, ensuring that the constant-velocity approximation remains valid over the short prediction horizon. 

\subsection{Trajectory optimization}
Given an initial skeletal path and a sequence of overlapping polyhedra from the SFC, the local planner computes a smooth, dynamically feasible trajectory for a quadrotor. The trajectory $\mathbf{p}(t): [0, T] \to \mathbb{R}^3$ is constrained to lie within the SFC. The total flight time $T$ is fixed and divided into $N$ segments with pre-assigned durations ${T_i}$. These durations come from the time assigned to ST-RRT$^*$ during path searching.

The optimization problem is formulated as a nonlinear minimization over the spatial path, leveraging the MINCO framework \cite{wang2022TRO} for efficient computation:

\begin{mini!}|s|[2]
{\mathbf{\xi}}{\mathcal{J}_s +  \mathcal{J}_d}{\label{eq:opt_full}}{}
\addConstraint{\mathbf{p}(t) = \C{M}(\B{q},\B{T}), \quad \forall t\in T_i,\ i=1,\dots,N}
\addConstraint{\mathbf{s}^{(0:2)}(0) = \mathbf{s}_0^{(0:2)}, \quad \mathbf{s}^{(0:2)}(T) = \mathbf{s}_f^{(0:2)}\label{eq:constraints}}
\end{mini!}

\noindent where $\C{J}_s$ is the squared-jerk control effort, $\C{J}_d$ is the penalty function that enforces dynamic limits like maximum velocity, maximum acceleration, etc. and corridor violation penalties. $\C{M}(\B{q},\B{T})$  is a linear-complexity smooth map from intermediate points $\B{q}$ and a time allocation $\B{T}$ (pre allocated based on ST-RRT$^*$) for all pieces to the coefficients of polynomial pieces. $\mathbf{\xi}$ denotes the vector whose elements encode the position of $\B{q}$, such that each element of $\B{q}$ can be represented as a convex composition of the vertices of corresponding polygon, with normalized and squared values of $\mathbf{\xi}$ ensuring that all points lie inside the polygon. The constraints in~\eqref{eq:constraints} corresponds to boundary conditions in position, velocity, and acceleration. 

The minimum jerk trajectory synthesized by MINCO is parametrized by only $\B{q}$ and $\B{T}$. The gradients of the objective function with respect to $\mathbf{\xi}$ are computed analytically, enabling efficient optimization using the L-BFGS algorithm. This formulation ensures the generated trajectory is safe (contained within the SFC), smooth, and respects the full dynamics of the quadrotor platform.

\subsection{Backup trajectory}

In this section, we introduce a backup planning algorithm to ensure reactive, collision-free flight when the UAV enters a deadlock or unsafe state. A deadlock is characterized by the failure of local path planning to progress toward the goal, thereby leaving the UAV exposed to potential collisions with dynamic obstacles.

The central principle of backup planning is to identify an intermediate safe goal in a direction away from obstacles, allowing the planner to reinitiate once the UAV reaches this region. The procedure consists of two stages:
\begin{enumerate}
    \item A convex polygon ($C_{static}$) is constructed around the UAV’s current location, incorporating only static obstacles. This guarantees that the backup maneuver is free from static collisions.
    \item The repulsive Cauchy Artificial Potential Field (CAPF) approach~\cite{srivastava2023ICUAS} is used to compute an escape direction vector pointing away from obstacles. The information about the size of the obstacles is taken from the bounding box dimensions recieved via the object detector. The intersection of this vector with $C_{static}$ defines the backup goal.
\end{enumerate}
A backup trajectory is then synthesized using ST-RRT$^\star$ with convex decomposition and trajectory optimization. The mechanism is triggered when either (i) a collision risk is detected along the current trajectory or (ii) the UAV is hovering while awaiting a feasible path. Once the UAV reaches the backup goal, the finite state machine (FSM) transitions back to its initial mode, thereby resuming its path planning toward the goal.

\section{Experiments and Validation}
\subsection{Simulation results}
\begin{figure}[t]
    \centering
    \begin{subfigure}[b]{0.4\linewidth}
        \centering
     \includegraphics[width=\linewidth]{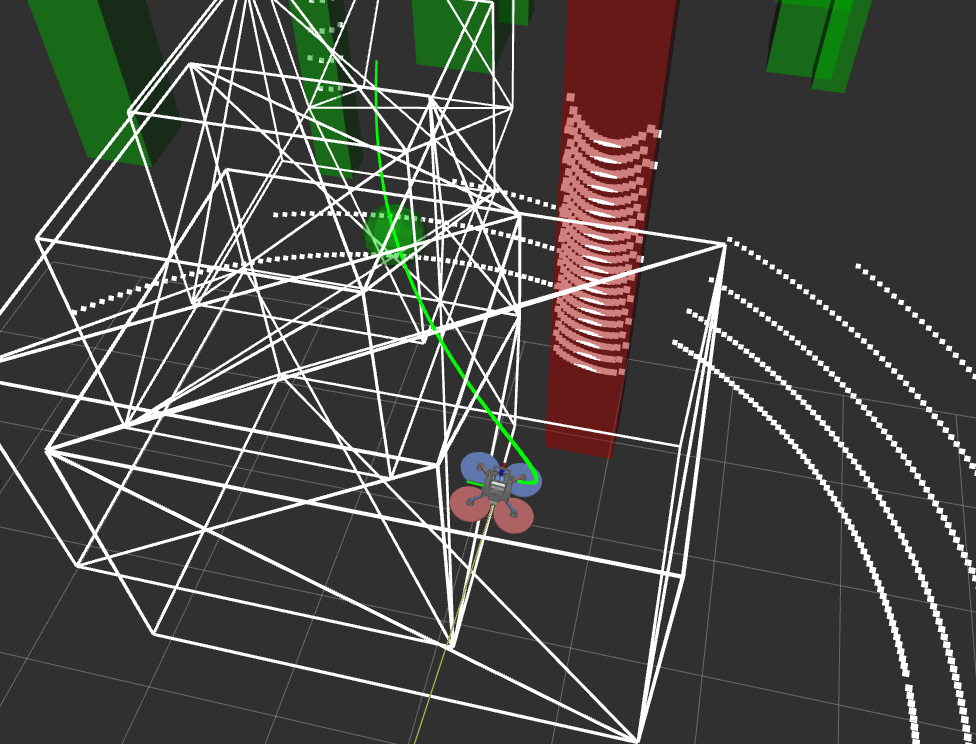}
        \caption{}
    \end{subfigure}
    \hfill
    \begin{subfigure}[b]{0.4\linewidth}
        \centering
        \includegraphics[width=\linewidth]{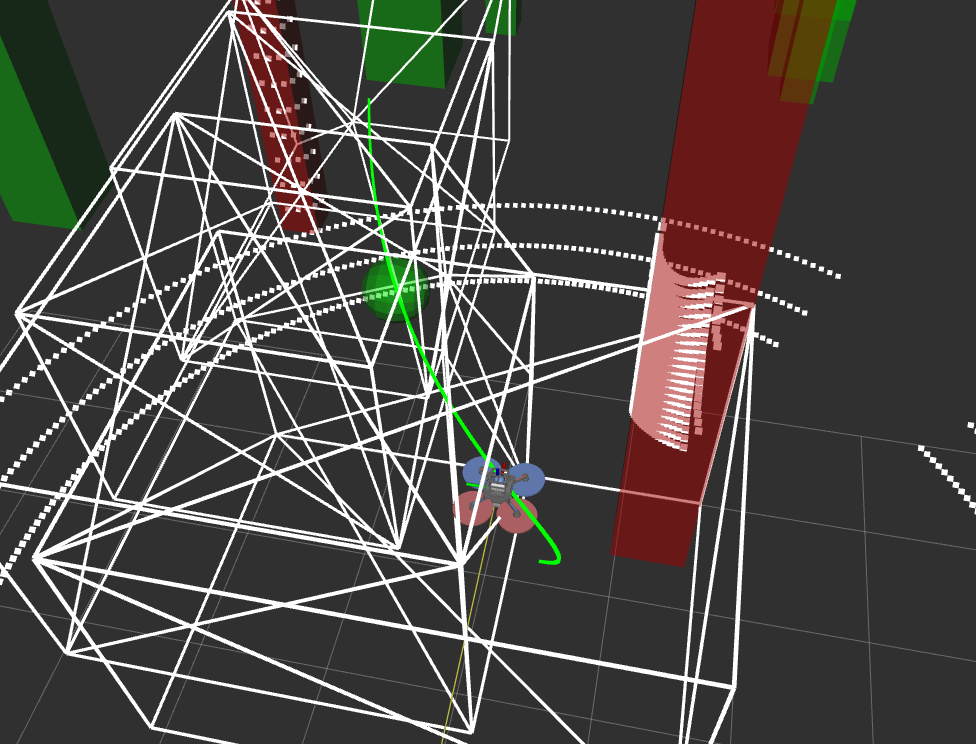}
        \caption{}
    \end{subfigure}
    
    \begin{subfigure}[b]{0.4\linewidth}
        \centering
        \includegraphics[width=\linewidth]{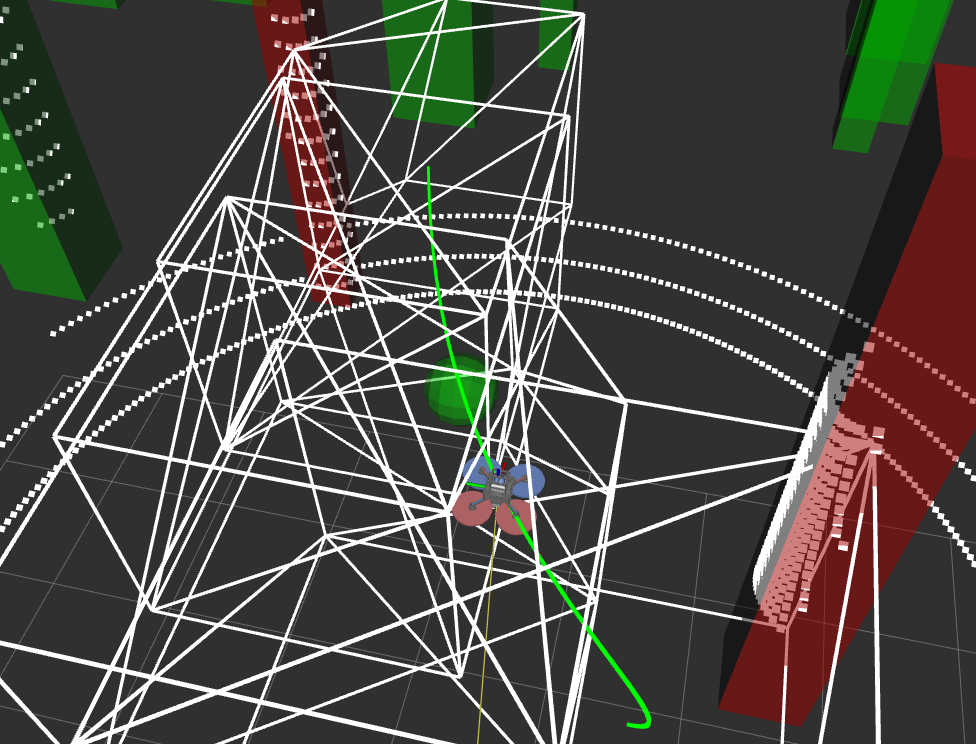}
        \caption{}
    \end{subfigure}
    \hfill
    \begin{subfigure}[b]{0.4\linewidth}
        \centering
        \includegraphics[width=\linewidth]{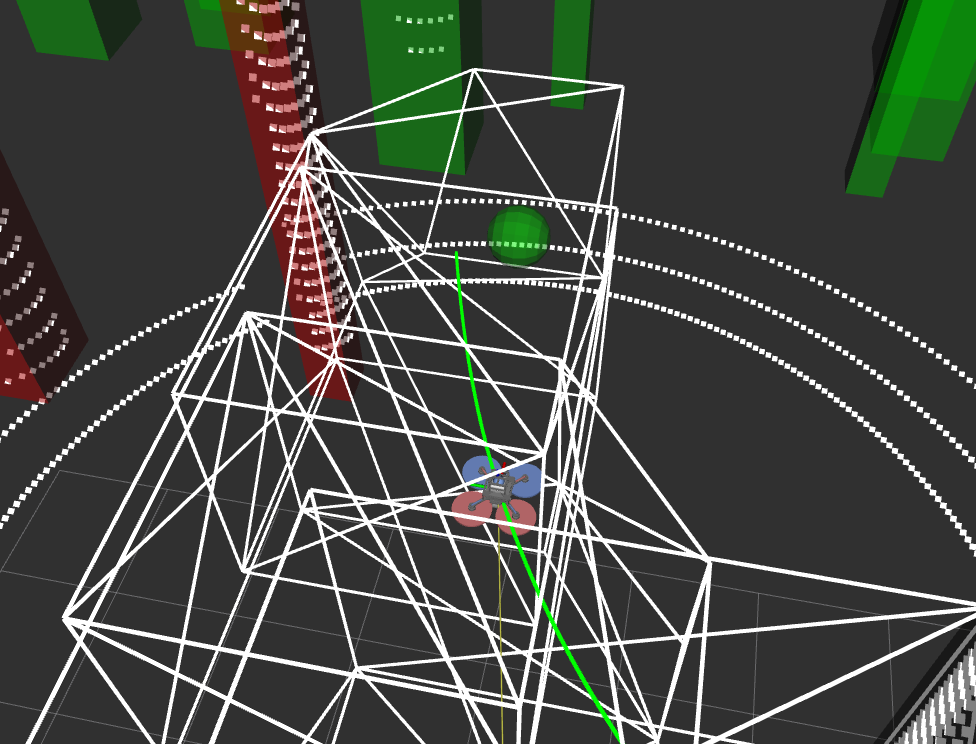}
        \caption{}
    \end{subfigure}
    
    \caption{Simulation of UAV avoiding dynamic obstacles. The white polygons represent the spatio-temporal free space around the UAV, which follows the green trajectory. The obstacles within red bounding boxes are with UAVs sensing range, while the obstacles represented as green cylinders lie outside of it.}
    \label{fig:drone_sim_avoid}
\end{figure}

In this section, we present simulation results for collision avoidance in dynamic, cluttered environments. The results are divided into two categories:
\begin{itemize}
    \item Simulation with ground truth obstacle information: In this setting, the ground-truth positions and velocities of obstacles within the UAV’s sensing range (6 m) are available, along with a 360$^\circ$ ground-truth point cloud. Simulations are performed in the \textit{gym-pybullet-drones} environment \cite{panerati2021learning} using a ROS~2 interface, and we do not assume perfect tracking conditions, rather relying on realistic simulation of inner loop flight control behaviour. We compare our algorithm against the methods in \cite{wu2024ICRA} and \cite{ren2025super} under identical conditions.
    \item Pipeline Validation: To evaluate the complete vision-based navigation pipeline and its potential for sim-to-real transfer, we conduct simulations in the PX4 SITL environment with Gazebo. In this case, obstacle information is obtained from an onboard detector~\cite{10323166}, rather than from ground-truth data.  
\end{itemize}
Fig.~\ref{fig:drone_sim_avoid} shows a snapshot from our simulation. All simulations were were executed on a workstation equipped with an Intel Core i7 (13th generation) processor and 16 GB of RAM.
\subsubsection{Ground-truth simulations}
The simulation environment is a $16 \times 16$ m$^2$ area. Dynamic obstacles are modeled as vertical cylinders with a height of 5 m and diameters uniformly sampled between 0.4 m and 1.0 m. Obstacle velocities are randomly sampled within the range 0–0.5 m/s. The maximum velocity of the UAV is capped at 1 m/s for our algorithm as well as for the method in~\cite{wu2024ICRA}. For~\cite{ren2025super}, we additionally report results at UAV speeds of 1 m/s and 5 m/s, since the approach is well suited for high-speed navigation. Simulations are conducted in three environments containing 10, 20, and 30 obstacles, respectively. For each environment, 50 independent trials are performed to evaluate the results. In addition, we compare our proposed approach, SPOT, against a variant of SPOT executed without the backup trajectory. The results are summarized in Table~\ref{tab:success_rate}.
 \begin{table}[t]
\centering
\caption{Comparison of success rates (\%) across methods for environments with increasing obstacle densities.}
\begin{tabular}{lccc}
\hline
\rule{0pt}{1.005\normalbaselineskip}
\textbf{Method} & \textbf{10 Obstacles} & \textbf{20 Obstacles} & \textbf{30 Obstacles} \\
\hline
\hline
\rule{0pt}{1.005\normalbaselineskip}
SPOT   & 100.0 & 92 & 80.2 \\
SPOT-w/o backup   & 94.2 & 82.4 & 52.2 \\
{\cite{wu2024ICRA}} & 100.0 & 71.3 & 57.1 \\
{\cite{ren2025super}-5m/s} & 100.0 & 70.0 & 62.2 \\
{\cite{ren2025super}-1m/s} & 52.2 & 42.8 &  10.2\\
\hline
\end{tabular}
\label{tab:success_rate}
\end{table}

\begin{table}[t]
\centering
\caption{Ablation of SPOT planner across obstacle densities}
\label{tab:spot_ablation}
\resizebox{\columnwidth}{!}{%
\begin{tabular}{c|cccc}
\hline
\textbf{Obstacles} & \textbf{Length (m)} & \textbf{Replans} & \textbf{Backup} & \textbf{Time (s)} \\
\hline
10 & 16.77 & 3 & 0 & 19.13 \\
20 & 23.73 & 11 & 2 & 34.56 \\
30 & 26.44 & 23 & 5 & 50.9 \\
\hline
\end{tabular}%
}
\end{table}

It is readily evident from the first two rows of Table~\ref{tab:success_rate} that incorporating a backup trajectory increases the success rate by more than 20\%. The improvement in the success rate of~\cite{ren2025super} at higher UAV velocities can be attributed to the comparatively low speed of obstacles. Specifically, when the UAV moves at 5 m/s, obstacle velocities correspond to at most 10\% of the UAV speed, compared to 50\% when the UAV moves at 1 m/s. This speed advantage enables more efficient navigation through the environment. However, even under these conditions,~\cite{ren2025super} performs worse than our approach due to the absence of object motion compensation in dynamic environments.

The primary reason for collisions in~\cite{wu2024ICRA} is its inability to escape deadlocks, that is, situations where no feasible path to the goal exists. In such cases, the UAV hovers at its current location, making it susceptible to collisions with dynamic obstacles. By contrast, our planner effectively avoids deadlocks by invoking the backup planner, as reflected in the higher success rates reported in Table~\ref{tab:success_rate}. An example of SPOT avoiding a deadlock and switching to a backup trajectory is illustrated in Fig.~\ref{teaser}.
In Table ~\ref{tab:spot_ablation} we provide mean ablations for the performance parameters of SPOT with different obstacle densities. In the table, Replans and Backup represent the number of times FSM had to switch to \textit{Initial} mode and \textit{Backup} mode respectively.
\subsubsection{Pipeline validation}
\begin{figure}[t]
    \centering
    \begin{subfigure}[b]{0.5\linewidth}
        \centering
        \includegraphics[width=\linewidth]{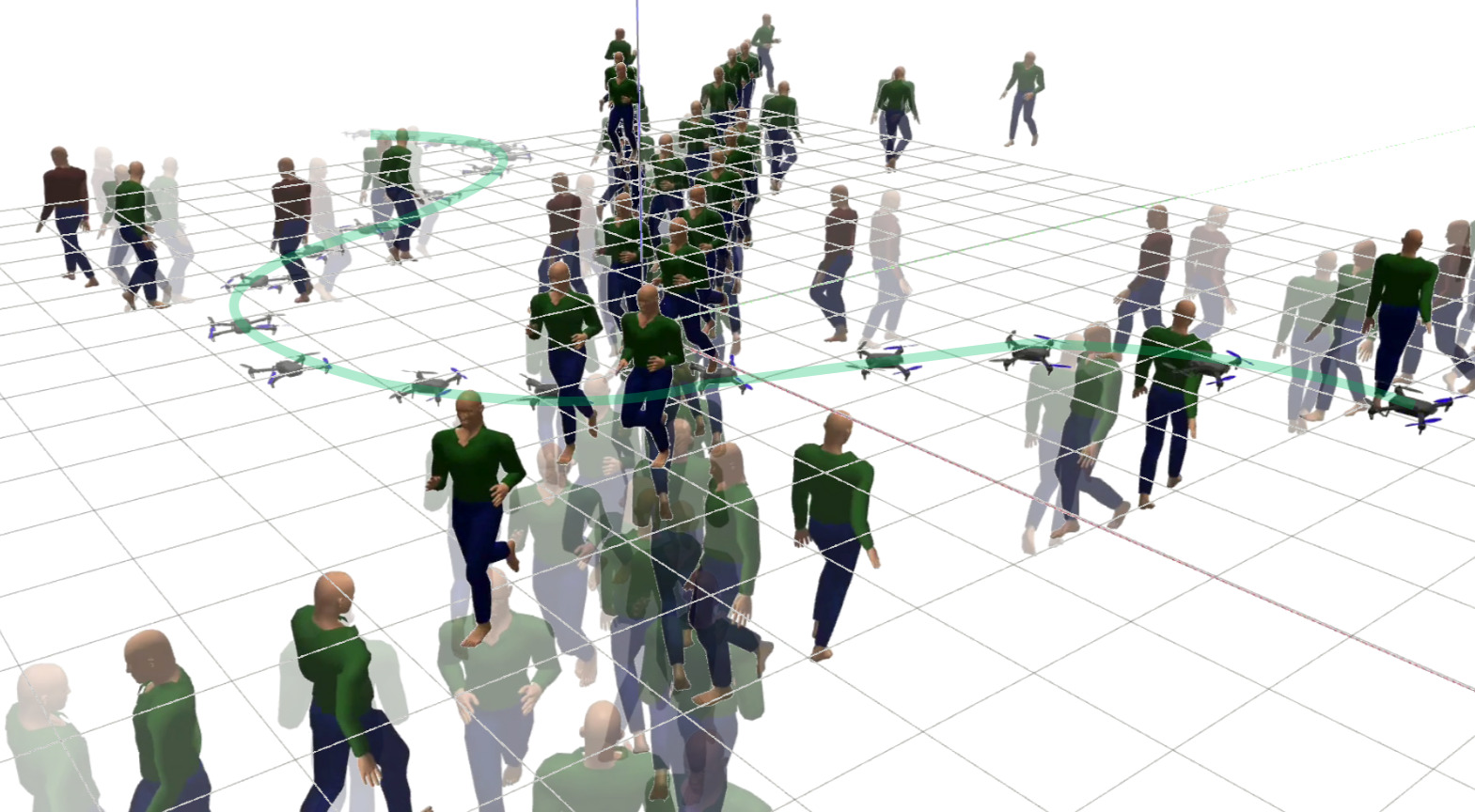}
        \caption{Sim-to-real validation in the PX4-Gazebo SITL environment for dynamic collision avoidance.}
    \end{subfigure}
    \begin{subfigure}[b]{0.5\linewidth}
        \centering
        \includegraphics[width=\linewidth]{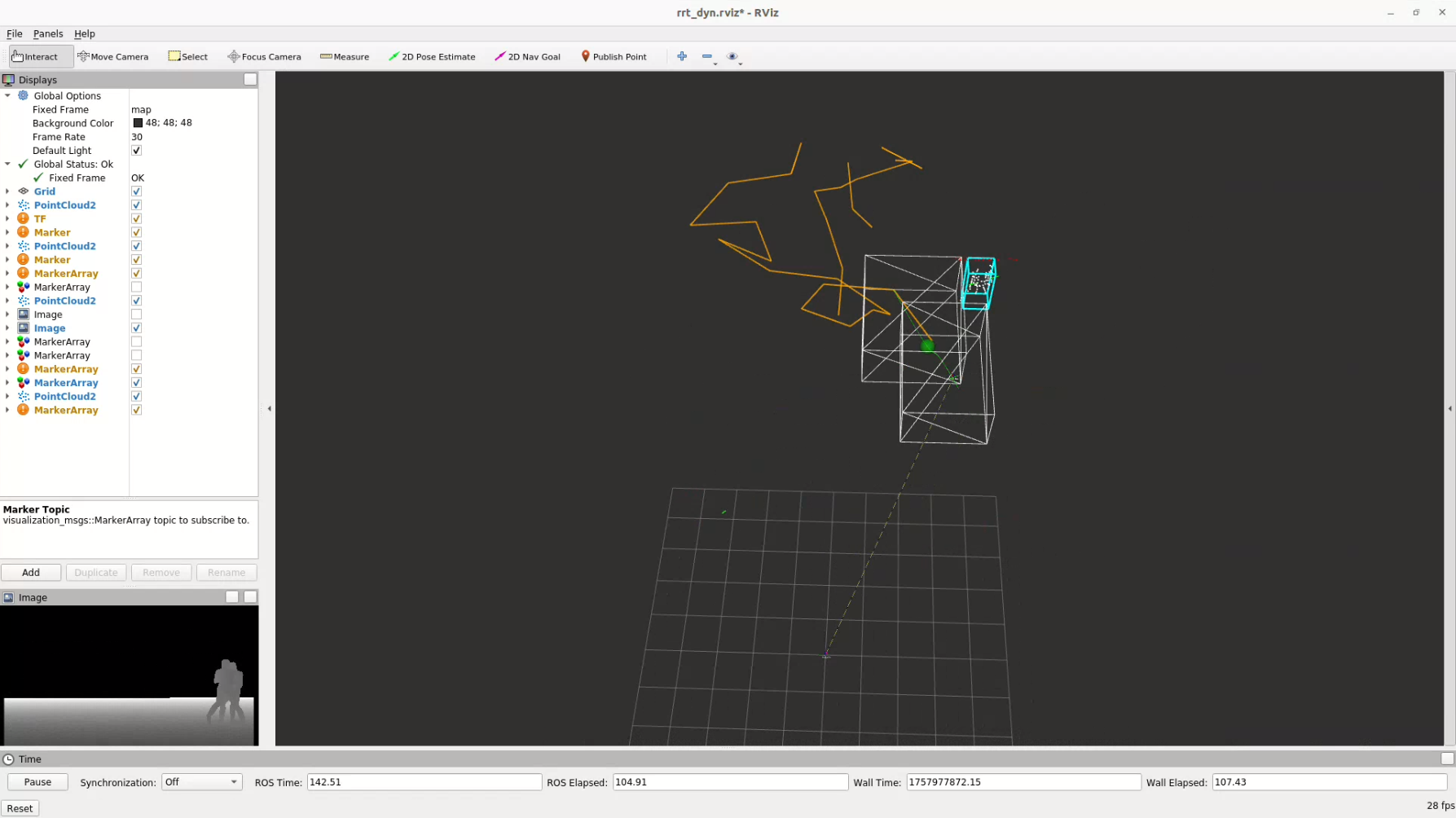}
        \caption{Onboard detector integrated with Spatio-Temporal Path planning}
    \end{subfigure}    
    \caption{Simulation of UAV avoiding humans with walking/running motion profiles. Bounding Boxes provided by onboard detector \cite{10323166} are used for segmenting pointclouds, which are then used by Spatio-Temporal planner for generating collision free trajectory}
\label{fig:gazebo_sim}
\end{figure}

To validate our sim-to-real pipeline for vision-based dynamic collision avoidance, we use the Gazebo simulation environment with PX4 SITL (see Fig.~\ref{fig:gazebo_sim}). Humans are simulated as dynamic obstacles, and the integration of our motion planning algorithm with the onboard detector is evaluated. Since the experiments use a depth camera, yaw planning is necessary to ensure the UAV faces the direction of motion while avoiding dynamic obstacles. For brevity, the yaw planning formulation is not presented here; readers can refer to~\cite{9422918} for details. The simulation is conducted with 5 humans traversing an area of 20*20 m$^2$. 

\subsection{Hardware Results}


\begin{figure}
    \centering
    \includegraphics[width=0.5\linewidth]{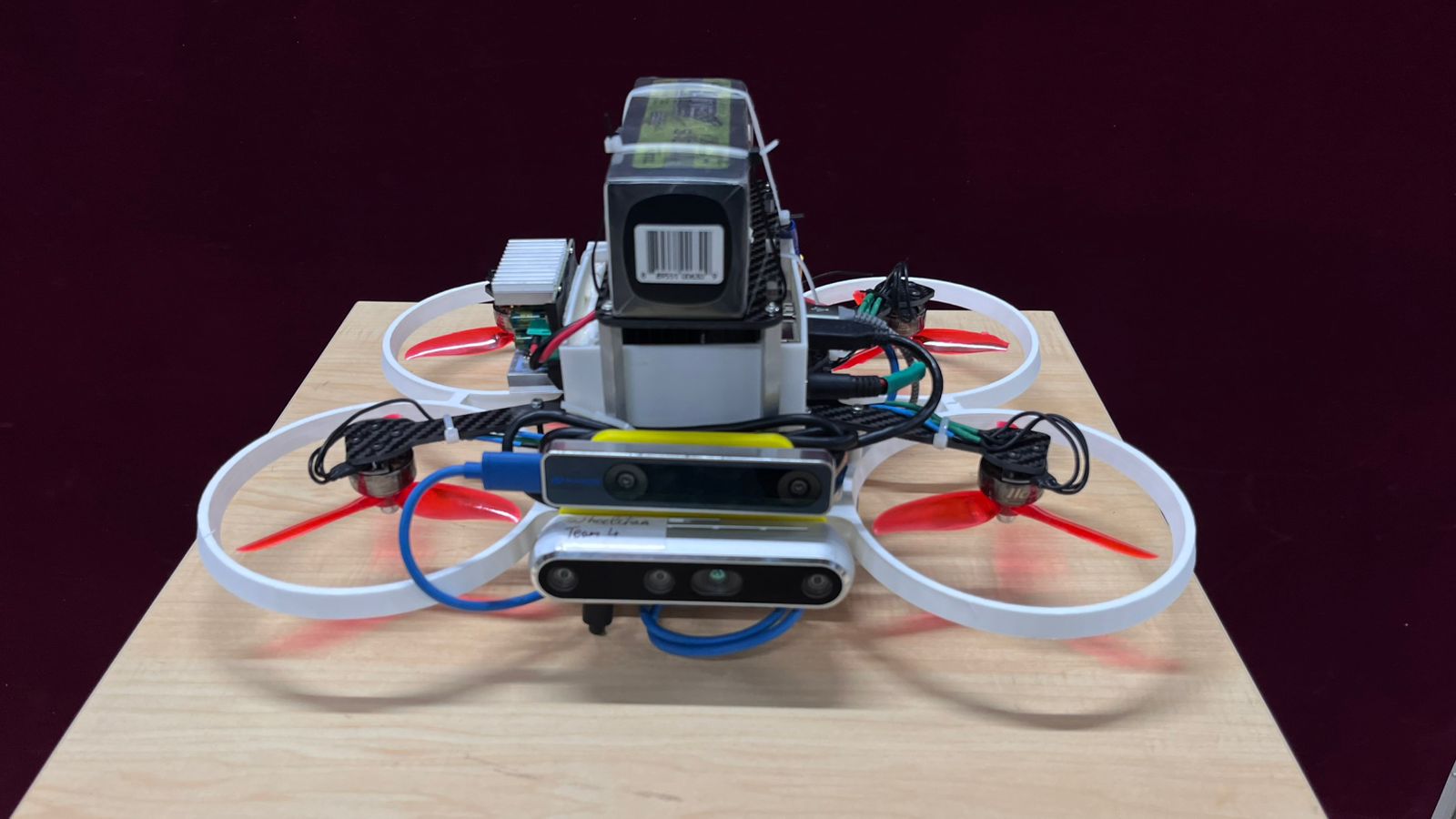}
    \caption{Hardware platform.}
    \label{fig:hardware_platform}
\end{figure}

We validate our collision avoidance algorithm through hardware experiments on an
in-house developed micro-quadcopter UAV. The platform, as shown in
Fig.~\ref{fig:hardware_platform}, has a wheelbase of 430\,mm and employs
5-inch, three-blade propellers for propulsion. For perception, the UAV is
equipped with an Intel RealSense D455, while state estimation is provided by an
Intel RealSense T265 camera for visual-inertial odometry. All onboard
computation is performed using an NVIDIA Jetson Xavier NX. The UAV runs PX4 on
an OmniNXT flight controller to enable trajectory tracking.

\begin{figure*}[t]
\centering
\subfloat[Start]{\includegraphics[trim=0 0 0 0,clip,scale=0.1]{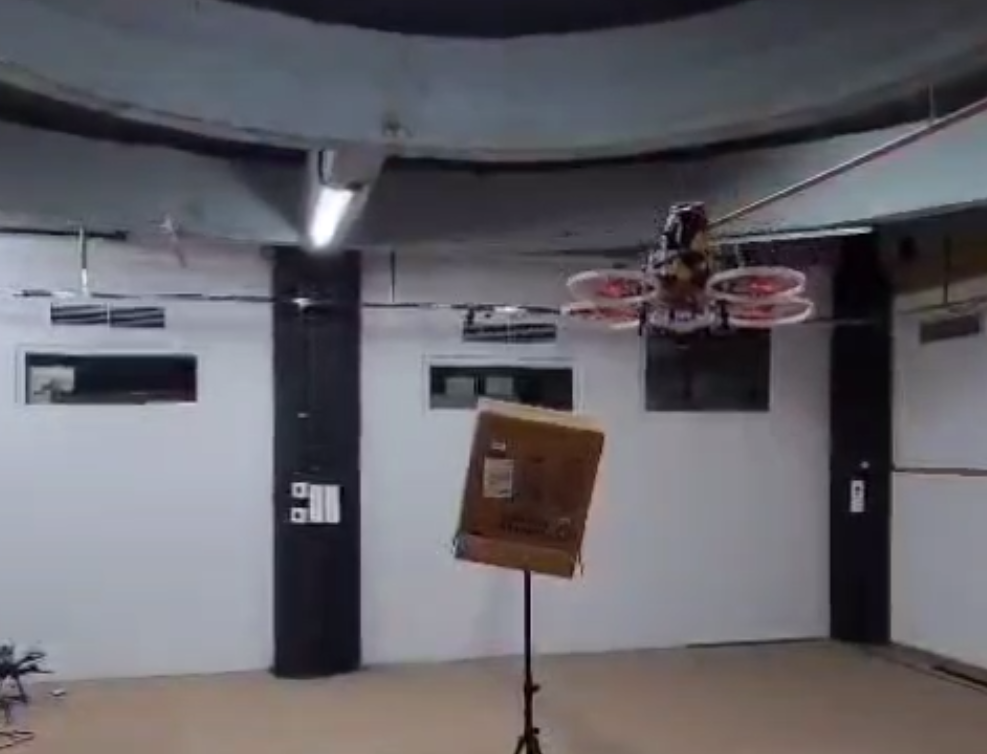}\label{fig:e1}}\hspace{0.01cm}
\subfloat[Approaching obstacle]{\includegraphics[trim=0 0 0 0,clip,scale=0.1]{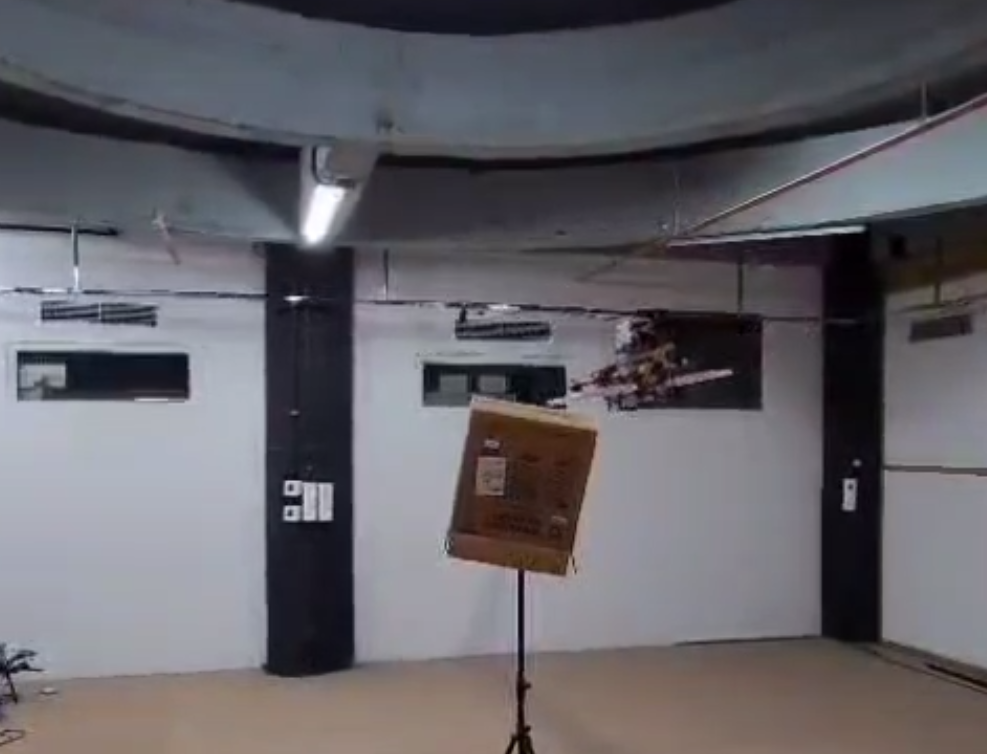}\label{fig:e2}}\hspace{0.01cm}
 \subfloat[Avoiding obstacle]{\includegraphics[trim=0 0 0 0,clip,scale=0.1]{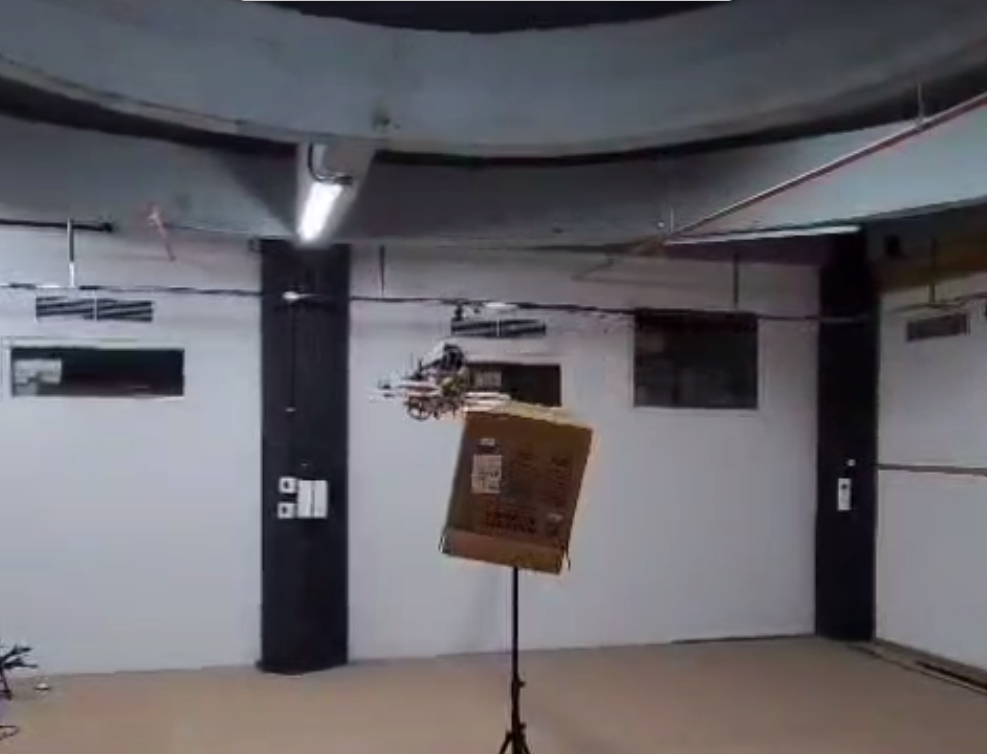}\label{fig:e3}}\hspace{0.01cm}
 \subfloat[Cleared obstacle]{\includegraphics[trim=0 0 0 0,clip,scale=0.1]{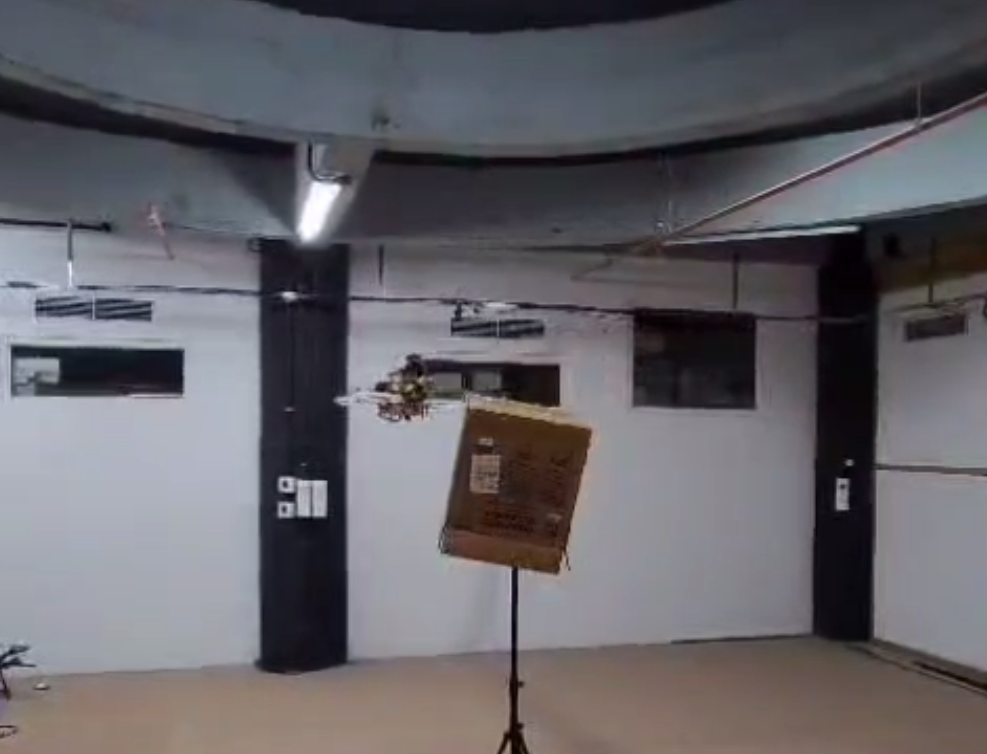}\label{fig:e4}}\hspace{0.01cm}
 \subfloat[Planned Trajectory]{\includegraphics[trim=0 0 0 0,clip,scale=0.097]{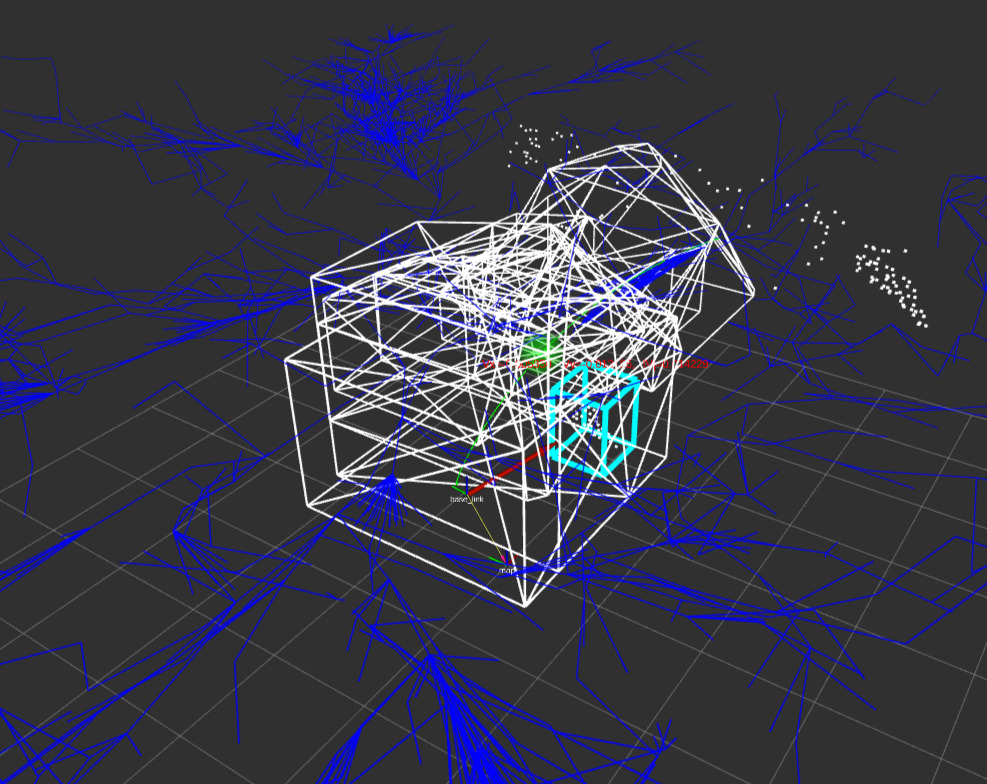}\label{fig:e5}}\hspace{0.01cm}\\
 \subfloat[Start]{\includegraphics[trim=0 0 0 0,clip,scale=0.1]{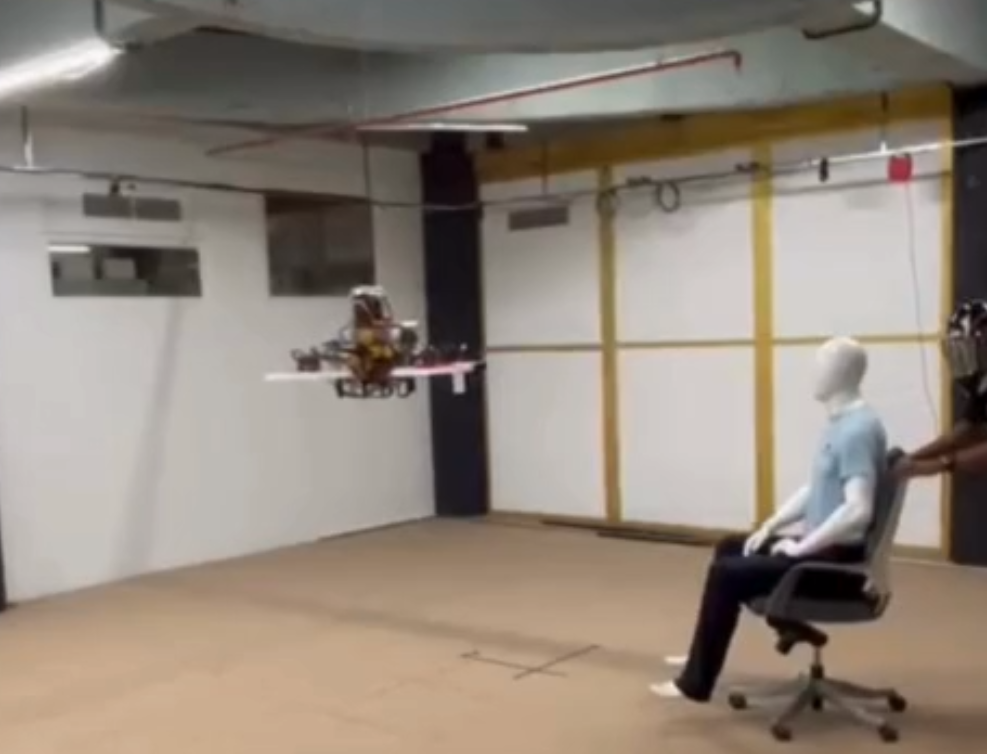}\label{fig:e5}}\hspace{0.01cm}
\subfloat[Approaching obstacles]{\includegraphics[trim=0 0 0 0,clip,scale=0.1]{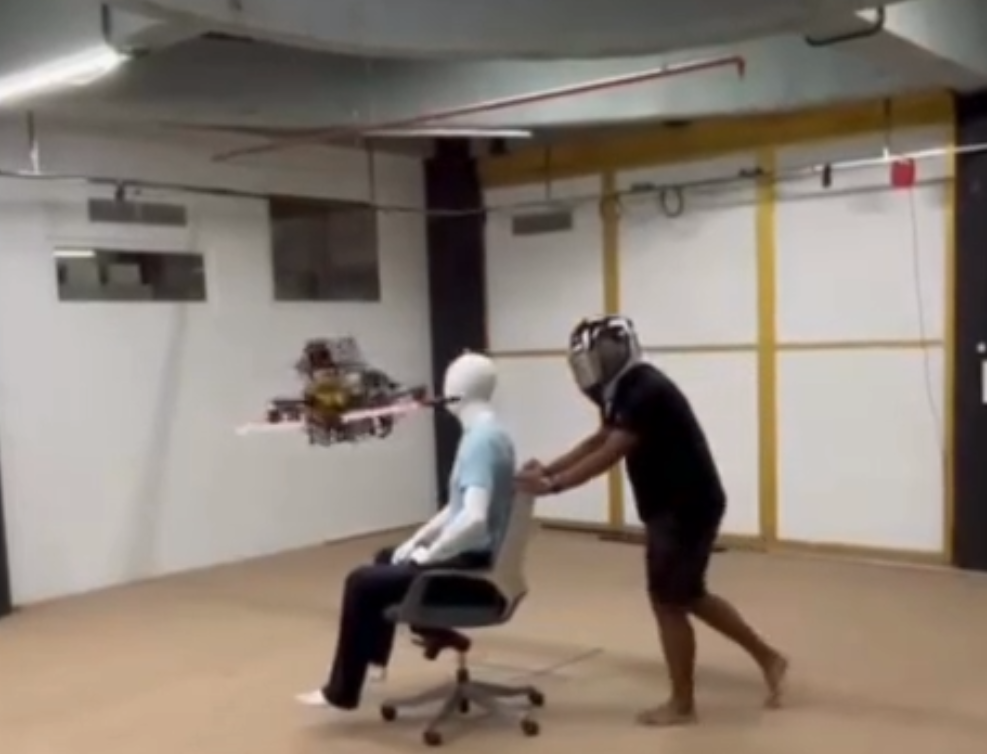}\label{fig:e6}}\hspace{0.01cm}
\subfloat[Avoiding obstacles]{\includegraphics[trim=0 0 0 0,clip,scale=0.1]{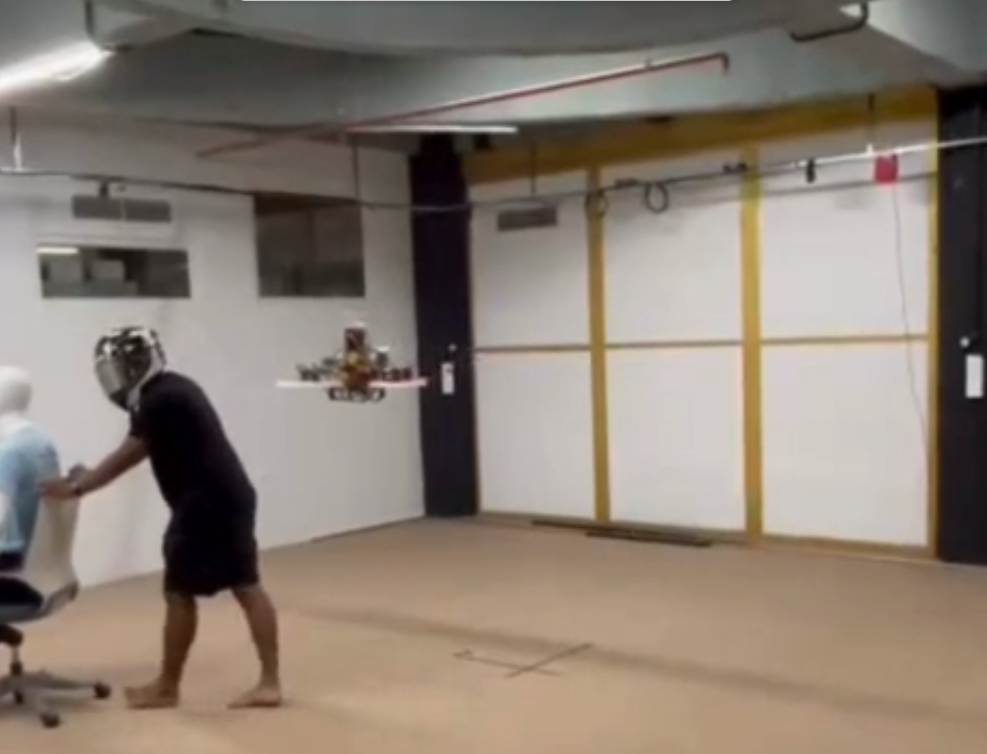}\label{fig:e7}}\hspace{0.01cm}
\subfloat[Cleared obstacles]{\includegraphics[trim=0 0 0 0,clip,scale=0.1]{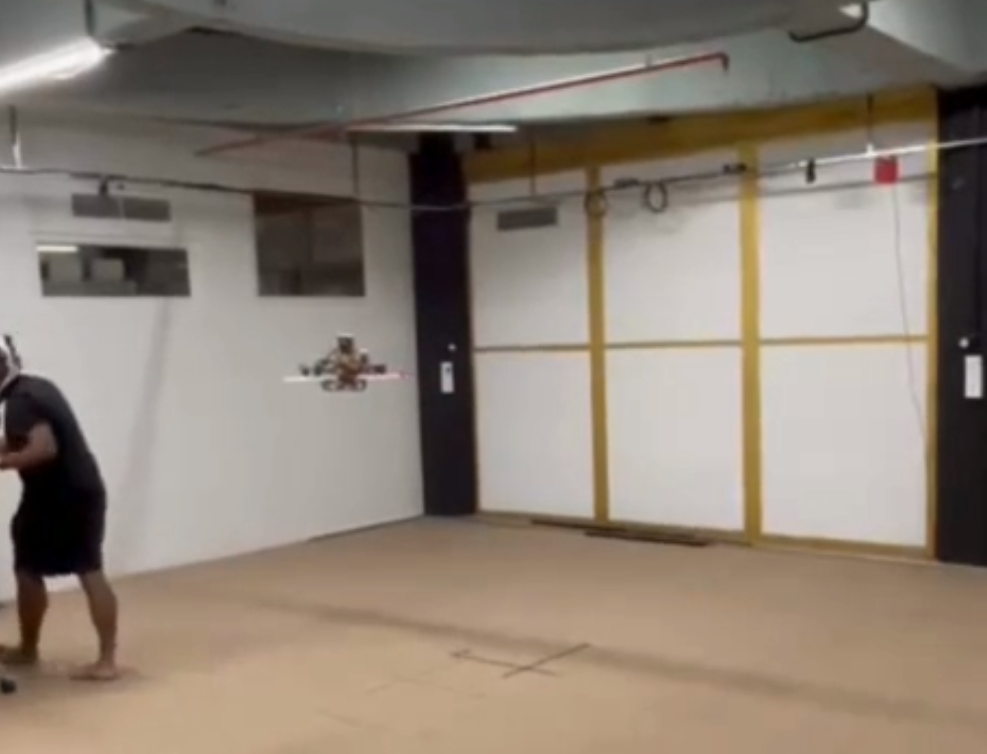}\label{fig:e8}}\hspace{0.01cm}
\subfloat[Planned Trajectory]{\includegraphics[trim=0 0 0 0,clip,scale=0.097]{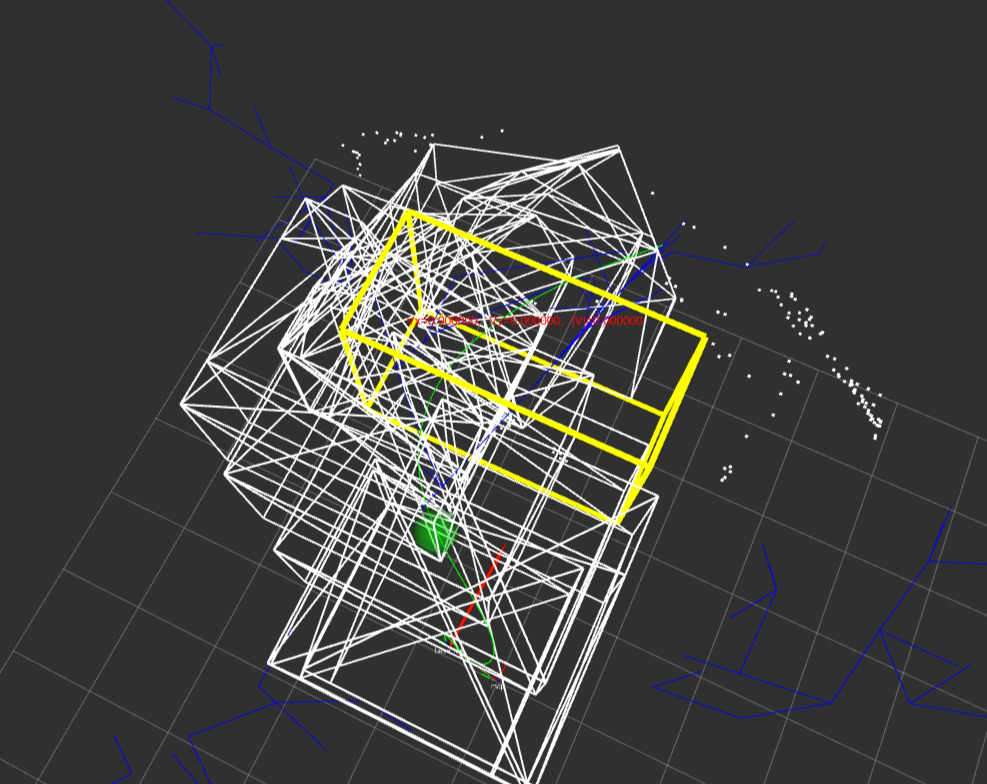}\label{fig:e5}}\hspace{0.01cm}
 \caption{Hardware experiments demonstrating collision avoidance on an in-house developed UAV. (a–d) UAV avoiding a static obstacle: starting from the initial position, approaching the obstacle, executing avoidance, and clearing it. (e) The planned trajectory. (f–i) UAV avoiding dynamic obstacles: starting from rest, approaching multiple moving obstacles, maneuvering around them, and successfully clearing the scene. (j) The planned trajectory. In the figures, white polygons represent obstacle free space while blue/yellow bboxes represent tracked obstacles}
  \label{fig:hardwar_experiments}
\end{figure*}

We conduct two sets of experiments to evaluate performance in the presence of static and dynamic obstacles, respectively. Unlike \cite{wu2024ICRA}, we use onboard computation to both track and avoid dynamic obstacles. Representative results are illustrated in Fig.~\ref{fig:hardwar_experiments}.

\section{Conclusion}
We presented a reactive motion planning framework for quadrotors in unknown, dynamic environments, combining a 4D Spatio-Temporal RRT planner with vision-based Safe Flight Corridor generation and trajectory optimization. Dynamic obstacles are detected and tracked using a vision-based pipeline, and a backup planning module ensures safe navigation in deadlock scenarios. Simulation and hardware experiments demonstrate the method’s performance and advantages over state-of-the-art approaches. Future work will focus on improving the estimation of the motion model for dynamic obstacles to enhance prediction and avoidance of collisions. We would also focus on improving the motion planning in more high speed and challenging environments.

\section{Acknowledgment}

The authors acknowledge the support provided by MeitY,
Govt. of India, under the project ”Capacity Building for
Human Resource Development in Unmanned Aircraft System
(Drone and Related Technology).”

\bibliography{citations.bib}
\bibliographystyle{IEEEtran}
\end{document}